\documentclass{article} 
\usepackage{iclr2026_conference,times}

\usepackage{amsmath,amsfonts,bm}

\def\eqref#1{equation~\ref{#1}}

\def\1{\bm{1}}

\DeclareMathAlphabet{\mathsfit}{\encodingdefault}{\sfdefault}{m}{sl}
\SetMathAlphabet{\mathsfit}{bold}{\encodingdefault}{\sfdefault}{bx}{n}

\usepackage{tabularx}
\usepackage{multirow}
\usepackage{booktabs}
\usepackage[table]{xcolor}
\usepackage[dvipsnames]{xcolor}
\usepackage{graphicx}
\usepackage{adjustbox}
\usepackage{booktabs}
\usepackage{subcaption}

\usepackage{hyperref}
\usepackage{url}
\usepackage{graphicx}
\usepackage{enumitem}

\usepackage{wrapfig}

\definecolor{CQColor}{rgb}{0.0,0.0,1.0} 

\definecolor{TSColor}{rgb}{0.5,0.0,0.8} 

\definecolor{CQRColor}{rgb}{1.0,0.0,0.0} 

\definecolor{mycolor_blue}{HTML}{E7EFFA}
\definecolor{mycolor_green}{HTML}{E6F8E0}
\definecolor{mycolor_gray}{HTML}{ECECEC}
\definecolor{pearDark}{HTML}{2980B9}
\definecolor{citecolor}{HTML}{2980b9}
\definecolor{linkcolor}{HTML}{c0392b}
\iclrfinalcopy

\title{Group Critical-token Policy Optimization for\\ Autoregressive Image Generation}

\author{Guohui Zhang\textsuperscript{1}, Hu Yu\textsuperscript{1}, Xiaoxiao Ma\textsuperscript{1}, Jinghao Zhang\textsuperscript{1,2}, Yaning Pan\textsuperscript{3}, Mingde Yao\textsuperscript{4}, \\ \textbf{Jie Xiao}\textsuperscript{1}, \textbf{Linjiang Huang}\textsuperscript{5}, \textbf{Feng Zhao}\textsuperscript{1\dag} \\
  \textsuperscript{1}University of Science and Technology of China, \textsuperscript{2}Shanghai Innovation Institute  \\
  \textsuperscript{3}Fudan University, \textsuperscript{4}CUHK, \textsuperscript{5}Beihang University
}

\iclrfinalcopy 
\begin{document}
{
  \renewcommand{\thefootnote}{\fnsymbol{footnote}}
  \footnotetext{\textsuperscript{\dag}Corresponding author.}
}
\maketitle

\begin{abstract}
Recent studies have extended Reinforcement Learning with Verifiable Rewards (RLVR) to autoregressive (AR) visual generation and achieved promising progress.
However, existing methods typically apply uniform optimization across all image tokens, while the varying contributions of different image tokens for RLVR's training remain unexplored. 
In fact, the key obstacle lies in how to identify more critical image tokens during AR generation and implement effective token-wise optimization for them. 
To tackle this challenge, we propose \textbf{G}roup \textbf{C}ritical-token \textbf{P}olicy \textbf{O}ptimization (\textbf{GCPO}), which facilitates effective policy optimization on critical tokens. 
We identify the critical tokens in RLVR-based AR generation from three perspectives, specifically: 
\textbf{(1)} Causal dependency: early tokens fundamentally determine the later tokens and final image effect due to unidirectional dependency;
\textbf{(2)} Entropy-induced spatial structure: tokens with high entropy gradients correspond to image structure and bridges distinct visual regions;
\textbf{(3)} RLVR-focused token diversity: tokens with low visual similarity across a group of sampled images contribute to richer token-level diversity. 
For these identified critical tokens, we further introduce a dynamic token-wise advantage weight to encourage exploration, based on confidence divergence between the policy model and reference model.
By leveraging 30\% of the image tokens, GCPO achieves better performance than GRPO with full tokens.
Extensive experiments on multiple text-to-image benchmarks for both AR models and unified multimodal models demonstrate the effectiveness of GCPO for AR visual generation. Code is available at \url{https://github.com/zghhui/GCPO}
\end{abstract}    
\section{Introduction}
\label{sec:intro}
Visual generative models~\citep{sun2024autoregressive, liu2024lumina, ma2024star} based on the autoregressive (AR) paradigm have made significant progress in the field of high-quality image generation.
Meanwhile, Reinforcement Learning (RL) with Verifiable Rewards (RLVR), demonstrated by OpenAI-o1~\citep{openaio1} and DeepSeek R1~\citep{guo2025deepseek} to enhance the reasoning abilities of large language models (LLM)~\citep{yang2025qwen3,team2025kimi}, is now being gradually introduced into the visual generation to improve preference alignment and task controllability. 

Recent works apply RLVR, especially Group Relative Policy Optimization (GRPO)~\citep{shao2024deepseekmath} for text-to-image generation by designing visual Chains-of-Thought (CoT)~\citep{jiang2025t2i}, optimizing reward functions~\citep{yuan2025ar}, and constructing customized datasets~\citep{pan2025focusdiff}.
Despite these advances, these methods typically assume that each token contributes equally to the RLVR's training objective and apply uniform policy optimization across the entire image token sequence.
While different tokens play distinct roles in text-to-image generation: some tokens determine and correspond to the global structure of the image, while others correspond to backgrounds or details.
Concurrent RLVR-based LLM reasoning works \citep{wang2025beyond,wang2025stabilizing} also realized such a functional distinction between tokens, and split them into reasoning-related \textit{critical tokens} and remaining knowledge-related tokens, where the former have higher entropy and dominate reasoning ability.
While this analogy highlights a shared imbalance in token importance, visual generation bears higher complexity due to the causal AR modeling and bidirectional image structure.

In this paper, we identify \textbf{critical tokens} in RLVR-based AR generation from three perspectives: 1) \textbf{Causal dependency} of AR; 2) \textbf{Entropy-induced spatial structure};  and 3) \textbf{RLVR-focused token diversity}. Specifically,
\textbf{1)} Early generated tokens continuously influence subsequent tokens and have a significant impact on the overall image structure due to the causal attention mechanism, as shown in Fig.~\ref{fig:initial_token_v2}.
\textbf{2)} For the token sequence of each image, we initially attempt to correlate entropy with image token, similar to the operations in \citep{wang2025beyond}. While, we find that the distribution of high/low entropy tokens didn't consistently correspond to certain parts of the image, like structure or background. Pushing one step further, we observe that the token entropy gradient map demonstrates a consistent spatial pattern among images, with the high value approximately corresponding to the structures and bridging distinct visual regions, see Fig.~\ref{fig:entropy_analyse}, which is sensitive to RL.
\textbf{3)} Within the group of GRPO, we observe that tokens deliver varying diversities for the same position along images. As shown in Fig.~\ref{fig:similarity}, tokens corresponding to background and texture regions tend to exhibit higher similarity, while tokens with lower similarity correspond to more complex region structures.

On the basis of our critical token selection strategy, we introduce \textbf{G}roup \textbf{C}ritical-token \textbf{P}olicy \textbf{O}ptimization (\textbf{GCPO}), a novel RLVR framework for AR image generation that facilitates effective policy optimization on critical image tokens. 
During each optimization step, GCPO first selects the critical tokens following the above three perspectives. Then, we further devise a dynamic token-level advantage weight for critical tokens to better encourage exploration. This dynamic weight is based on confidence divergence of critical tokens between the updating policy model and reference model, which differs from the standard GRPO that allocates advantage uniformly for each token. Finally, we only retain the policy gradients of critical tokens to perform policy optimization.
By utilizing only critical tokens (\textbf{30\%} of the tokens), GCPO achieves better performance than GRPO with full tokens on multiple text-to-image generation benchmarks. Extensive experiments demonstrate the effectiveness of GCPO, including Geneval, T2I-CompBench, and Human Preference Benchmark, which is also verified on both AR models and unified multimodal models. In summary, our contributions are as follows:
\begin{itemize}
    \item We identify \textbf{critical tokens} in RLVR-based AR visual generation from three perspectives: Causal dependency of AR, Entropy-induced spatial structure, and RLVR-focused token diversity, achieving a structure-centric and comprehensive critical-token selection strategy.

    \item We propose \textbf{Group Critical-token Policy Optimization (GCPO)}, a RLVR framework for autoregressive image generation that facilitates effective policy optimization on critical image tokens. We further devise the dynamic advantage weight strategy for critical tokens to enable reasonable exploration and stabilize the inherently generative prior, based on their confidence divergence between the policy model and the reference model.

    \item GCPO is applicable to both AR models and unified multimodal models. Extensive experiments demonstrate that GCPO, by optimizing only critical tokens (\textbf{30\%} of the tokens), achieves better performance than GRPO with full tokens across multiple T2I benchmarks. 
\end{itemize}

\section{Related Work}
\label{sec:Related Work}

\textbf{Autoregressive Visual Generation}. Autoregressive image generation models~\citep{sun2024autoregressive, liu2024lumina, xie2024show, team2024chameleon,ma2024star} adopt the next token prediction paradigm, which has been widely applied in large language models, to enable text-to-image generation. Representative works such as DALL-E~\citep{ramesh2022hierarchical}, LlamaGen~\citep{sun2024autoregressive}, and Lumina-mGPT~\citep{liu2024lumina} typically first use an image tokenizer~\citep{esser2021taming} to discretize continuous image data into a sequence of tokens, and then employ decoder-only transformer architecture to model visual tokens, thereby achieving high-quality image synthesis. Furthermore, recent research has focused on unifying image generation and image understanding within a single architecture. These models are capable of accepting diverse types of input (e.g., text, image, video) and producing one or more modalities as output. 
For instance, 
Janus-Pro~\citep{chen2025janus} adopts a dual-encoder structure that separately processes textual and visual data. Transfusion~\citep{zhou2024transfusion}, Show-o~\citep{xie2024show}, and BAGEL~\citep{deng2025emerging} further combine the strengths of Transformer and Diffusion architectures for multimodal understanding and generation, achieving superior performance across various tasks.

\textbf{Reinforcement Learning for Visual Generation}. Reinforcement Learning with Verifiable Rewards (RLVR) has achieved significant progress in the field of large language models (LLMs). A series of open-source models (e.g., DeepSeek-R1~\citep{guo2025deepseek} and Kimi K1.5~\citep{team2025kimi}) and RLVR methods~\citep{yu2025dapo, yue2025vapo,shrivastava2025sample} have been proposed, further advancing the development of this field.
Meanwhile, recent efforts~\citep{liu2025flow,xue2025dancegrpo,pan2025focusdiff} have increasingly explored the potential of RLVR, especially Group Relative Policy Optimization~\citep{shao2024deepseekmath} (GRPO), in the field of visual generation. SimpleAR~\citep{wang2025simplear} has demonstrated that GRPO can significantly enhance the aesthetic quality and prompt alignment of AR models. T2i-R1~\citep{jiang2025t2i} leverages GRPO to jointly optimize both semantic-level and token-level Chain-of-Thought (CoT) reasoning processes, thereby improving the generative capabilities of a unified multimodal model. In addition, Focus-Diff~\citep{pan2025focusdiff} constructs a fine-grained dataset for better capturing and distinguishing fine-grained semantic differences. ReasonGen-R1~\citep{zhang2025reasongen} further proposes an improved adaptive entropy loss function to effectively mitigate the issue of mode collapse. 

\textbf{Reinforcement Learning for Critical tokens}. Recently, several studies~\citep{li2025llms, vassoyan2025ignore} have focused on the deeper analysis and exploration of the role of RL in LLMs' reasoning task, especially at the token level. Critical Tokens Matter~\citep{lin2024critical} suggests that identifying and replacing critical tokens can significantly improve the model's accuracy, and proposes a contrastive estimation method to accurately locate these tokens. ConfPO~\citep{yoonconfpo} further investigates the effectiveness of selectively optimizing only low-confidence and information-rich tokens. Additionally, \citep{wang2025beyond, wang2025stabilizing} points out the existence of ``fork" tokens in LLM reasoning paths, indicating that these tokens typically have high entropy and are related to logical reasoning. They further observe that these fork tokens (critical tokens) can be identified by entropy and play more critical role than other tokens in enhancing the LLM's reasoning ability. However, identifying critical tokens in AR visual generation and implementing effective token-wise optimization for them remains unexplored.
\section{Preliminary}
\label{sec:Preliminary}
\textbf{Autoregressive Image Generation.}  
A common autoregressive (AR) model includes two main components: an image tokenizer and an AR transformer~\citep{esser2021taming,yu2021vector}. For image tokenizer, typically VQ-VAE~\citep{van2017neural}, it converts images $\mathcal{I}\in\mathcal{R}^{H\times W\times3}$ into discrete tokens sequence $Z = \{z_1, \ldots, z_{N}\}$, where each token $z_t \in \mathcal{V}$. $\mathcal{V}$ and $t$ represent the VQ-VAE codebook and token index in sequence, respectively. Next, the AR model autoregressively predicts the joint distribution of the next image token conditioned on the text and previously generated tokens: $P(z|c) = \prod_{t=1}^{N} P(z_t | z_{<t}, c)$, where $c$ represents text embedding.

\textbf{Token Entropy.} The AR model outputs probability distribution over the codebook $\mathcal{V}$ for each token. Therefore, we can use the following formula to calculate the entropy~\citep{shannon1948mathematical} of this distribution, which is referred to as the token entropy $H$:
\begin{equation}
    H(z_t) = -\sum_{k=1}^{V} P_{t,k} \log P_{t,k}
\end{equation}

\textbf{Group Relative Policy Optimization (GRPO).} For each prompt $p$, GRPO~\citep{shao2024deepseekmath} samples a group of image outputs $\{o^1, o^2, \ldots\}$ from the old policy model $\pi_{\theta_{old}}$. Then, it computes the corresponding rewards $\{r^1, r^2, \ldots\}$ for each output within each group. The advantage $A^i$ is calculated from the rewards of the group, and each output in the group shares the same advantage. Then optimizes the policy model $\pi_{\theta_{old}}$ by maximizing the following objective:
\begin{equation}
\begin{aligned}
\mathcal{J}_{\text{GRPO}}(\theta)  &= 
\mathbb{E}_{\{o^i\}_{i=1}^G \sim \pi_{\theta_{\text{old}}}}  \\
\Bigg[ \frac{1}{G} &\sum_{i=1}^{G} \frac{1}{|o^i|}\sum_{t=1}^{|o^i|}
\color{black} \Bigg( \min\Big(r^i_{t}(\theta)\hat{A}^i,\,
    \text{clip}\left(r^i_{t}(\theta), 1 - \varepsilon, 1 + \varepsilon\right) \hat{A}^i \Big)
   - \beta {D}_{\text{KL}}(\pi_\theta \| \pi_{\text{ref}})
\Bigg)
\Bigg],
\end{aligned}
\end{equation}
where 
\begin{equation}
    A^i = \frac{r^i - mean(\{r^1, r^2, \cdots, r^G\})}{std(\{r^1, r^2, \cdots, r^G\})}, \quad
    r^i_{t}(\theta) = \frac{
    \pi_\theta(o^i_{t} \mid q, o^i_{<t})
}{
    \pi_{\theta_{\text{old}}}(o^i_{t} \mid q, o^i_{<t})
},
\label{con:grpo}
\end{equation}

\section{Observation and Analysis in AR Visual Generation}
\label{sec:Analysis}
\subsection{Causal dependency of AR}
\label{sec:AR dependency}
\begin{wrapfigure}{r}{0.3\textwidth} 
    \centering 
    \vspace{-\intextsep} 
    \includegraphics[width=\linewidth]{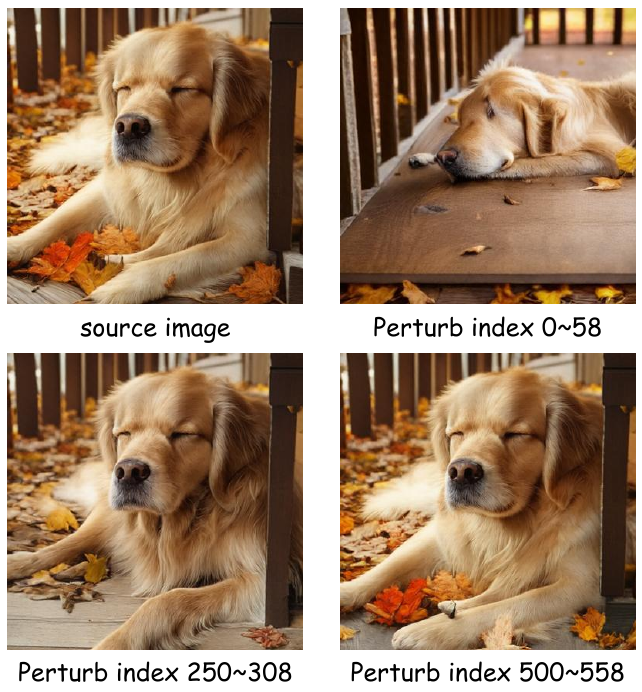}
    \caption{Visual results of perturbing different tokens.} 
    \label{fig:initial_token_v2} 
    \vspace{-\intextsep} 
\end{wrapfigure}
The core capability of AR models stems from the causal attention mechanism in transformers. Under this next token prediction paradigm, the initially generated tokens continuously influence the generation of all subsequent tokens, thereby significantly impacting the overall structure and layout of the image. To further validate this point, we inject additional noise into the tokens at different positions during the generation process~\citep{beyer2025highly}. This causal influence is visibly illustrated in Fig.~\ref{fig:initial_token_v2}, where perturbations to the early 58 tokens (token index from 0 to 58) introduce substantial changes to the image's global structure, while perturbations to the middle 58 tokens (token index from 250 to 308) only affect local details. This empirical evidence confirms that early tokens serve as global priors and structural guides. In contrast, later tokens are constrained by both preceding tokens and local consistency, making them more focused on generating local content and details. Therefore, the initial tokens should be part of critical tokens, as these early decisions propagate throughout the entire AR generation and establish the foundation for high-quality visual structures.

\subsection{Entropy-induced spatial structure}
\label{sec:Image Spatial Structure}
Concurrent RLVR-based LLM reasoning works~\citep{wang2025beyond,wang2025stabilizing} have found a functional distinction between text tokens, and split them into reasoning-related critical tokens and knowledge-related tokens based on \textit{token entropy}. The former with high entropy (such as ``wait", ``however") serve as logical \textbf{connectors} that bridge consecutive reasoning parts in CoT, while the latter with low entropy primarily capture factual or domain-specific knowledge.

\begin{figure}[h]
    \centering
    \includegraphics[width=0.98\linewidth]{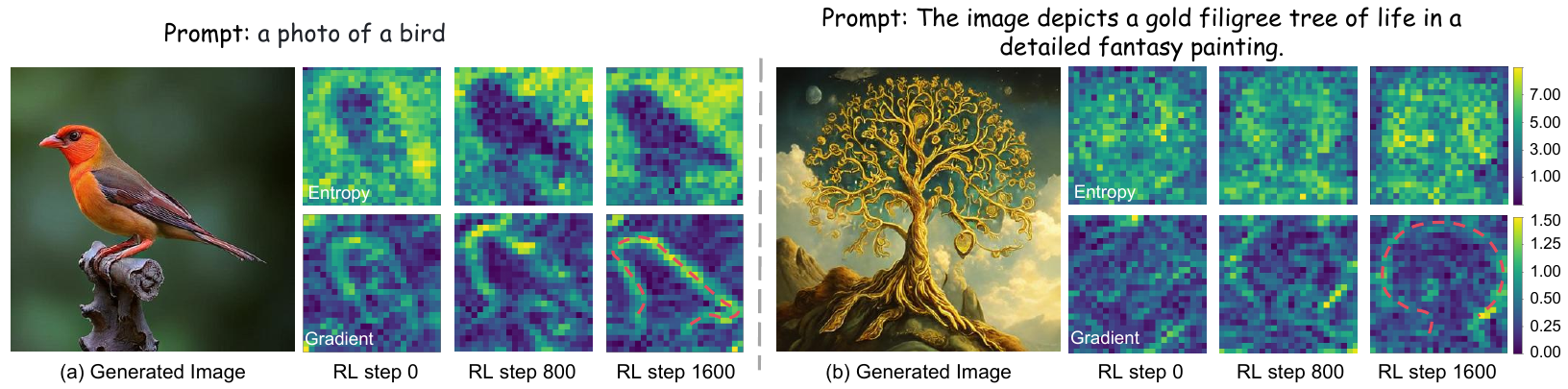}
    \caption{The entropy maps of images exhibit distinct spatial patterns. Tokens with high entropy gradient consistently correspond to image structure, which gradually strengthens with RL training.}
    \label{fig:entropy_analyse}
\end{figure}

Inspired by this, we first analyze the entropy distribution of image tokens. As shown in Fig.~\ref{fig:entropy_analyse}, we observe that the distribution of high/low entropy tokens corresponds to different parts of the image in different prompts: tokens with high entropy mainly correspond to the background in Fig.~\ref{fig:entropy_analyse} (a), while corresponding to the subject in Fig.~\ref{fig:entropy_analyse} (b). This phenomenon gradually becomes obvious in RL training. We argue that it is likely influenced by image and prompt complexity, and we provide more analysis results in Sec.~\ref{sec:appendix_entropy_analysis}. Furthermore, we find that this entropy distribution in images exhibits a regional and spatial pattern, where tokens within the subject or background regions have approximate entropy values. We argue that image tokens exhibit strong spatial locality, with neighboring pixels sharing similar visual characteristics and entropy values~\citep{he2024zipar,xiang2025make}.

Pushing one step further, we analyze the 2D gradient of entropy and observe consistent spatial patterns among different images. Specifically, tokens with high entropy gradients typically correspond to subject structure or regions with significant structural variation, and this pattern becomes more pronounced with RL training. These tokens exhibit large entropy changes in neighboring tokens, serving as \textbf{connectors} to link distinct visual regions. Based on this insight, we point out that tokens with high entropy gradients are critical and sensitive for spatial structures, and entropy gradients can serve as a universal and reliable proxy for identifying tokens associated with image structure.

\subsection{RLVR-focused token diversity}
\label{sec:RL-based visual property}

\begin{wrapfigure}{r}{0.4\textwidth} 
    \centering 
    \vspace{-\intextsep} 
    \includegraphics[width=\linewidth]{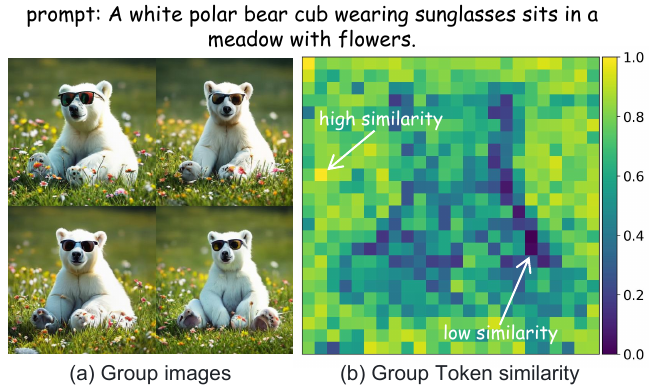}
    \caption{The average cosine similarity of tokens at corresponding location across a group of images.} 
    \label{fig:similarity} 
    \vspace{-\intextsep} 
\end{wrapfigure}
GRPO typically relies on sample-level reward signals, where the differences between samples guide the direction of policy optimization. A group of similar samples provides limited reward information, which restricts the model's performance improvement and training efficiency. In light of this,~\citep{he2025rewarding,yu2025dapo,chen2025dra} focus on enhancing sample diversity through entropy regularization or encouraging semantic diversity. In AR visual generation task, we further focus on token-level diversity. Within the group of GRPO, we observe that tokens deliver varying diversities for the same position along images. As shown in Fig.~\ref{fig:similarity}, tokens in background and texture regions tend to exhibit higher similarity and hardly reflect the visual differences among images. While tokens with lower similarity correspond to more complex regional structures and contribute to richer information for GRPO-based policy optimization. Therefore, we select tokens with low similarity as part of critical tokens.

\subsection{Dynamic advantage weight}
\label{sec:Dynamic advantage weight}
\begin{wrapfigure}{r}{0.4\textwidth} 
    \centering 
    \vspace{-\intextsep} 
    \includegraphics[width=\linewidth]{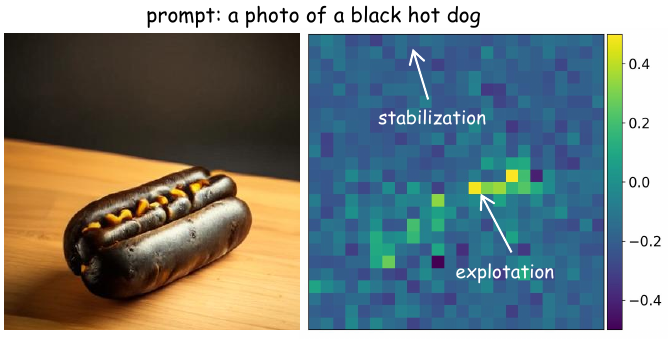}
    \caption{Confidence divergence between the Policy and Reference model} 
    \label{fig:confidence} 
    \vspace{-\intextsep} 
\end{wrapfigure}
Balancing exploration with stabilization of generative priors is crucial for different tokens during RLVR training~\citep{wang2025stabilizing}. We argue that different critical tokens should have dynamic and distinct exploration constraints. Specifically, initial tokens should explore more moderately to prevent global structural collapse, while tokens with high entropy gradients and low similarity should have stronger exploration. Considering that the reference model itself serves as the starting point for training, we analyze the confidence divergence of each token between the training policy model and the reference model. As shown in Fig.~\ref{fig:confidence}, we observe that this divergence is not only dynamic as the policy model updating, but also has distinct constraints: initial tokens exhibit smaller confidence divergence, while tokens corresponding to structures show larger divergence. 
Based on this, we utilize this dynamic divergence as weight to eliminate complex manual specification.

\begin{figure}[t]
    \centering
    \includegraphics[width=\linewidth]{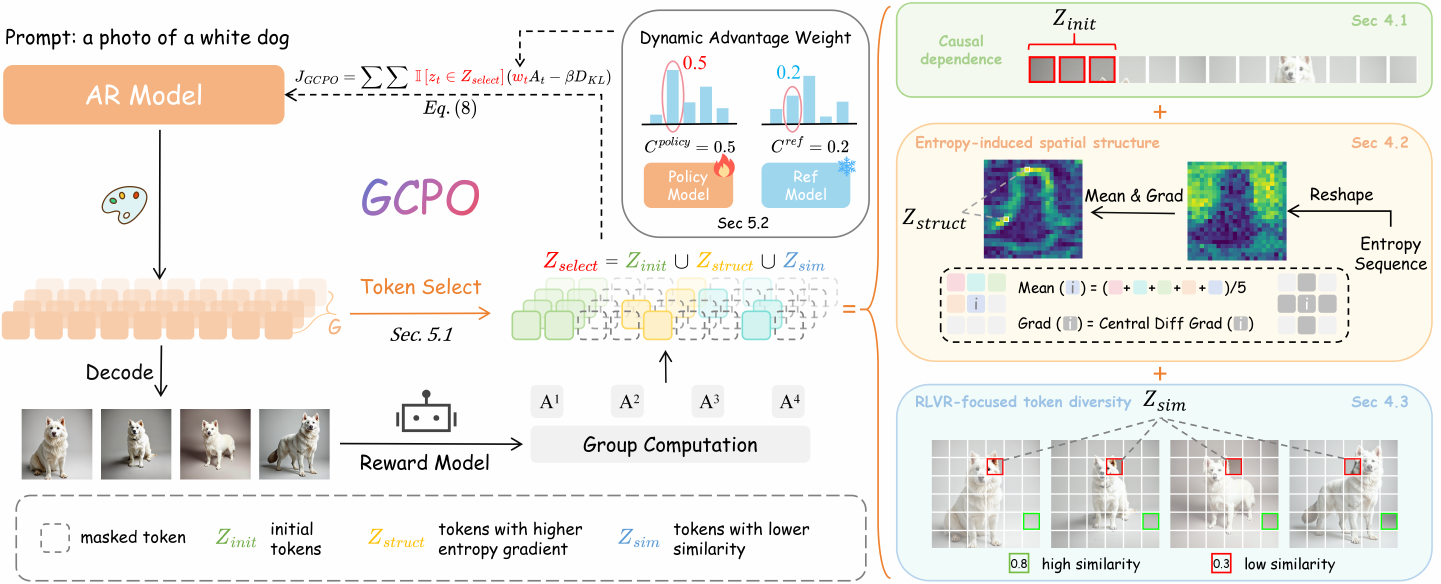}
    \caption{\textbf{Overview of GCPO.} GCPO first generates a group of images for each prompt and obtains the corresponding reward. For token selection, we first select the initial token as $Z_{init}$. Then, we calculate the local mean and central difference gradient in 2D entropy map, and select the token with the higher gradients as $Z_{struct}$. Subsequently, we select the tokens with lower similarity as $Z_{smi}$ based on intra-group cosine similarity at each position. For dynamic advantage weight, we calculate the cumulative mean confidence difference of each token as the advantage weight $w_{t}$. Finally, we only retain the policy gradients of critical tokens $Z_{select}$ to perform policy optimization.}
    \label{fig:framework}
\end{figure}

\section{Group Critical-token Policy Optimization}
\label{sec:Methology}
\subsection{Token Selection}
The Sec.~\ref{sec:Analysis} discusses the critical image tokens selection from three aspects.  Here, we detail the token selection strategy, as illustrated in Fig.~\ref{fig:framework}. Specifically, we denote the image token sequence as $Z = \{z_1, \ldots, z_{N}\}$, where $N$ is the total length of the sequence. The initial image tokens is represented as $Z_{init} = \{z_1, \ldots, z_{K_{init}}\}$, with $K_{init}$ indicating the number of initial tokens. 

Subsequently, we reshape the entropy sequence $\{H_t\}^N_{t=1}$ associated with these tokens into a 2D entropy map to select structure-related tokens. To mitigate the influence of noise in the entropy map, we perform a local averaging operation. Considering the local spatial and causal dependency of image tokens, the averaging is performed as follows:
\begin{equation}
\bar{H_t} = mean(H_{t} + H^{(l,u)}_{t} + H^{(u)}_{t} + H^{(r,u)}_{t} + H^{(l)}_{t}),
\end{equation}
where $(l,u), (u), (r,u)$ and $(l)$ denote the upper-left, upper, upper-right and left neighboring positions of $H_t$, respectively. Then, we use the central difference to calculate the gradient of each token in the average entropy map. We select the $K_{struct}$ tokens with the largest gradients as $Z_{struct}$.

Next, we calculate the cosine similarity of token embeddings at each sequence position within group of images. Specifically, for each token sequence position $t$, we consider the group of token embeddings $\{e_{t,1}, e_{t,2}, \ldots, e_{t,G}\}$ derived from $G$ images. The pairwise cosine similarity between token embeddings at position $t$ is calculated as follows:
\begin{equation}
S^{(t)}_{jk} = \frac{e_{t,j} \cdot e_{t,k}}{\|e_{t,j}\| \|e_{t,k}\|}, \quad 1 \leq j < k \leq G
\end{equation}
We then calculate the average pairwise similarity $\bar{S}_t$ for each sequence position $t$. We select the $K_\text{similarity}$ tokens with the lowest average based on $\bar{S}$. 

Finally, the overall critical token selection set is defined as the union of the three subsets:
\begin{equation}
Z_{select} = Z_{init} \cup Z_{struct} \cup Z_{sim}
\end{equation}
By default, the size of each subset $K_{init}$, $K_{struct}$, and $K_{sim}$ is set to 10\% of the total token sequence length, ensuring balanced and representative selection.

\subsection{Dynamic Advantage Weight}

The Sec.~\ref{sec:Dynamic advantage weight} discusses the motivation of the dynamic advantage weight. Furthermore, considering that the token at position $t$ is predicted by its preceding tokens, we further employ the cumulative average of confidence divergence as the weight for each critical token. This can be formalized as:
\begin{equation}
w_t = \frac{1}{t} \sum_{j=1}^{t} clip \left(C^{policy}_j - C^{ref}_j, -\epsilon_w, \epsilon_w \right)
\end{equation}
where $w_i$ denotes the advantage weight at position $t$. $C^{policy}_j$ and $C^{ref}_j$ represent the confidence (model's log probability) of the $j$-th token on the policy model and the reference model, respectively. $\epsilon_w$ is the clip coefficient to prevent excessive weights from influencing training stability.

\begin{figure}[htbp]
    \centering
    \includegraphics[width=\linewidth]{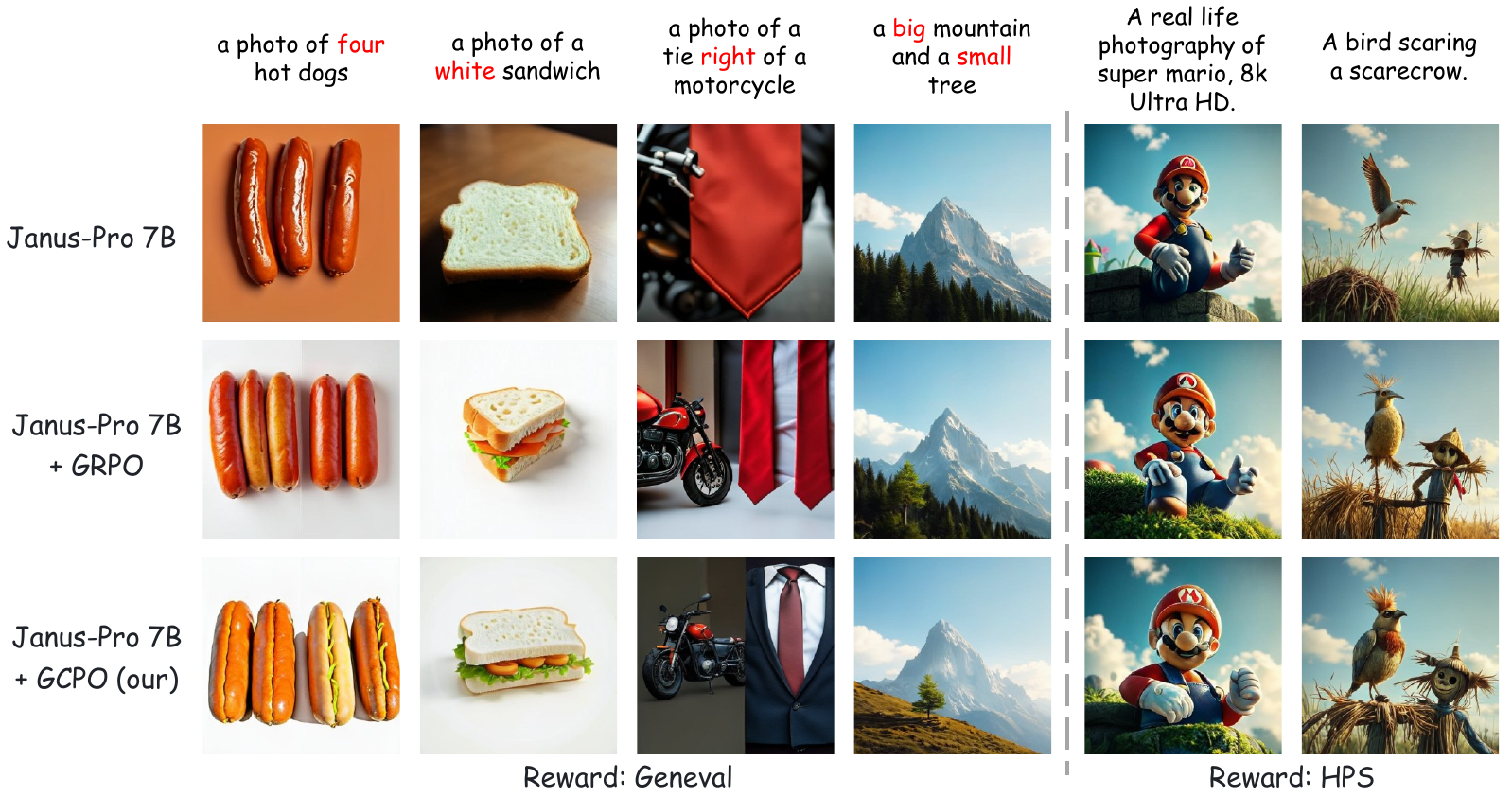}
    \caption{Visualization Results. We provide image generation results on Counting, Color, Position, and Shape tasks, as well as image quality.}
    \label{fig:visual_compare}
\end{figure}

\subsection{Objective Function}
The overall objective of GCPO is formulated as follows:
\begin{equation}
\begin{aligned}
\mathcal{J}_{\text{GRPO}}(\theta) = 
\mathbb{E}_{\{o^i\}_{i=1}^G \sim \pi_{\theta_{\text{old}}}}  
\Bigg[ \frac{1}{G} \sum_{i=1}^{G} &\frac{1}{|o^i|}\sum_{t=1}^{|o^i|} \color{red}\mathbb{I}\left[ z_{t} \in Z_{select}   \right]
\color{black} \Bigg(
    \color{red}w^i_{t}\color{black} \min\Big(r^i_{t}(\theta)\hat{A}^i,\,\\
   &  \text{clip}\left(r^i_{t}(\theta), 1 - \varepsilon, 1 + \varepsilon\right) \hat{A}^i \Big)
   - \beta \mathbb{D}_{\text{KL}}(\pi_\theta \| \pi_{\text{ref}})
\Bigg)
\Bigg],
\end{aligned}
\label{con:gcpo}
\end{equation}
where $\color{red}\mathbb{I}[\cdot]$ is the indicator function that evaluates to 1 if the condition inside holds and 0 otherwise.

The differences are highlighted in red between Equation (~\ref{con:gcpo}) and Equation (~\ref{con:grpo}): \textbf{(i)} The optimization term of each token is multiplied by $\mathbb{I}\left[ z_{t} \in Z_{select} \right]$, ensuring that only critical tokens in $Z_{select}$ are involved in the overall optimization objective; \textbf{(ii)} The advantage term of each critical token is multiplied by $w_t$, allocating token-wise advantage weight to encourage exploration. 

\section{Experiments}
\label{sec:Experiments}

In this section, we evaluate GCPO to improve the performance of AR model and the unified multimodal model on two representative tasks. (1) Composition Image Generation: We report the results on GenEval and T2i-Compbench, which primarily evaluate the models’ ability on spatial relationships and color attributes.  (2) Image quality and Human Preference Alignment: We report DEQA, ImageReward, and PickScore metrics, reflecting the visual quality and human preference of images. 

\subsection{Experimental Setup}
\textbf{Training Settings.} For the composition image generation task, our training dataset is sourced from 50,000 training prompts generated by Geneval pipeline, following~\cite{liu2025flow}. For the image quality and human preference alignment task, we utilize 15000 prompts from the HPSv2 training dataset. We conduct evaluations on LlamaGen~\citep{sun2024autoregressive}, Janus-Pro 1B~\citep{chen2025janus}, and Janus-Pro 7B~\citep{chen2025janus}. Please refer to the Appendix~\ref{sec:Detailed Experimental Setup} for detailed training settings. 

\textbf{Benchmark.} We evaluate our method on Geneval~\citep{ghosh2023geneval}, T2I-CompBench~\citep{huang2023t2i}, and DrawBench~\citep{saharia2022photorealistic} to comprehensively validate effectiveness. 
T2I-CompBench is a comprehensive benchmark for open-world compositional text-to-image generation, covering six compositional task categories.
DrawBench contains comprehensive and challenging prompts designed to assess the generative capabilities of T2I models. We use it to evaluate  DEQA-Score~\cite{you2025teaching}, ImageReward~\citep{xu2023imagereward} and Pick Score~\citep{kirstain2023pick} metrics. We also report HPSv2 score~\citep{wu2023human} on HPSv2 Benchmark test data.

\textbf{Reward Model.} For the composition image generation task, following~\citep{liu2025flow}, we adopt Geneval reward as our reward model. For the image quality and human preference alignment task, we use HPSv2 as our reward model. Notably, given the relatively lower performance of LlamaGen on Geneval, we only use HPSv2 as the reward model for this base.
\begin{table}[htbp]
\centering
\caption{ \label{tab:generation_geneval} Quantitative comparison results on the GenEval benchmark. The best result is in \colorbox{mycolor_green}{green}.}
\resizebox{\linewidth}{!}{
\begin{tabular}{c|ccccccc}
    \toprule
    \textbf{Method} & \textbf{Overall}$\uparrow$ & \textbf{Sing Obj.}$\uparrow$ & \textbf{Two Obj.}$\uparrow$ & \textbf{Counting}$\uparrow$ & \textbf{Color}$\uparrow$ & \textbf{Position}$\uparrow$ & \textbf{Color Attr.}$\uparrow$ \\
    \midrule
    \multicolumn{8}{c}{\textit{Diffusion-based Method}} \\
    \hline
    PixArt-$\alpha$~\citep{chen2024pixart}   & 0.48  & 0.98 & 0.50 & 0.44 & 0.80 & 0.08 & 0.07 \\ 
    SDXL~\citep{podell2023sdxl} & 0.55 & 0.98 & 0.74 & 0.39 & 0.85 & 0.15 & 0.23 \\
    SD3~\citep{esser2024scaling} &  0.63 & 0.98 & 0.78 & 0.50 & 0.81 & 0.24 & 0.52 \\  
    DALL-E 3~\citep{betker2023improving} & 0.67 & 0.96 & 0.87 & 0.47 & 0.83 & 0.43 & 0.45 \\
    FLUX.1-dev~\citep{Flux} & 0.66 & 0.98 & 0.81 & 0.74 & 0.79 & 0.22 & 0.45 \\
    \hline
    \multicolumn{8}{c}{\textit{AR-based method}} \\
    \hline
    LlamaGen~\citep{sun2024autoregressive} & 0.32 & 0.71 & 0.34 & 0.21 & 0.58 & 0.07 & 0.04 \\
    Emu3~\citep{wang2024emu3}  & 0.54 & 0.98 & 0.71 & 0.34 & 0.81 & 0.17 & 0.21  \\
    Show-o~\citep{xie2024show}& 0.68 & 0.98 & 0.80 & 0.66 & 0.84 & 0.31 & 0.50 \\
    GPT-4o~\citep{chatgpt4o} & 0.85 & 0.99 & 0.92 & 0.85 & 0.91 & 0.75 & 0.66 \\
    
    Janus-Pro-1B~\citep{chen2025janus} & 0.73 & 0.98 & 0.82 & 0.51 & 0.89 & 0.65 & 0.56 \\
    Janus-Pro-7B~\citep{chen2025janus} & 0.80 & 0.99 & 0.89 & 0.59 & 0.90 & 0.79 & 0.66 \\
    \hline
    \multicolumn{8}{c}{\textit{AR-based Method + RL}} \\
    \hline
    Show-o+PARM~\citep{guo2025can} & 0.69 & 0.97 & 0.75 & 0.60 & 0.83 & 0.54 & 0.53 \\
    T2I-R1~\citep{jiang2025t2i} & 0.79 & 0.99 & 0.91 & 0.53 & 0.91 & 0.76 & 0.65 \\
    LlamaGen+GRPO & 0.39  & 0.83  & 0.41  & 0.28  & 0.68  & 0.11  & 0.06  \\ 
    Janus-Pro-1B+GRPO & 0.84 & \colorbox{mycolor_green}{1.00} & 0.95 & 0.59 & 0.84 & 0.88 & \colorbox{mycolor_green}{0.77} \\
    Janus-Pro-7B+GRPO & 0.87 &	0.99 &	0.92 &	0.71 &	\colorbox{mycolor_green}{0.94} &	0.92 &	0.73  \\
    \hline
    \multicolumn{8}{c}{\textit{Our}} \\
    \hline
    LlamaGen+GCPO & 0.42   & 0.83   & 0.49   & 0.25   & 0.71   & 0.13  & 0.08  \\            
    Janus-Pro-1B+GCPO & 0.85 & \colorbox{mycolor_green}{1.00} & \colorbox{mycolor_green}{0.96} & 0.63 & 0.88 & 0.91 & 0.73 \\
    Janus-Pro-7B+GCPO & \colorbox{mycolor_green}{0.90} &	0.99 &	0.95 &	\colorbox{mycolor_green}{0.90} &	0.90 &	\colorbox{mycolor_green}{0.95} &	0.76  \\
    \bottomrule
\end{tabular}
}
\end{table}
\begin{table}[ht]
\centering
\caption{ \label{tab:t2icompbench_quality} Comparison results on T2I-CompBench and DrawBench, evaluated by DEQA-Score, ImageReward, and PickScore. The best result is in \colorbox{mycolor_green}{green}.}
\renewcommand{\arraystretch}{1.25}
\setlength{\tabcolsep}{1.15mm}{
\resizebox{0.99\linewidth}{!}{
\begin{tabular}{c|cccccc|ccccc}
    \toprule
    \textbf{Method} 
    & \textbf{Color}$\uparrow$ & \textbf{Shape}$\uparrow$ & \textbf{Texture}$\uparrow$ & \textbf{Spatial}$\uparrow$ & \textbf{Non-Spat.}$\uparrow$ & \textbf{Complex}$\uparrow$
    & \textbf{DEQA}$\uparrow$ & \textbf{HPS}$\uparrow$ & \textbf{ImgRwd}$\uparrow$ & \textbf{PickScore}$\uparrow$ \\

    \midrule
    \multicolumn{11}{c}{\textit{AR-based method}} \\
    \hline
    LlamaGen~\citep{sun2024autoregressive} & 0.4202 	& 0.3967 &	0.5103 &	0.0772 &	0.3050 &	0.2908 & 2.70 &21.62   & -0.36 & 20.27  \\            
    Janus-Pro-1B~\citep{chen2025janus} & 0.3439 & 0.2363 & 0.2788  & 0.0969 & 0.2813 & 0.2733  & 3.38 & 25.46 & -0.19 & 20.79  \\
    Janus-Pro-7B~\citep{chen2025janus} & 0.6359 & 0.3528 & 0.4936  & 0.2061 & 0.3085 & 0.3559  & 3.55 & 28.00 & 0.68  & 21.82   \\
    \hline
    \multicolumn{11}{c}{\textit{AR-based Method + RL}} \\
    \hline
    LlamaGen+GRPO & 0.4454 &	0.4092 &	0.5446 &	0.0780 &	0.3069 	& 0.3066 & 2.85 & 25.28 & -0.14 & 20.44  \\ 
    Janus-Pro-1B+GRPO & 0.7050 & 0.3150 & 0.4621  & 0.3020 & 0.2963 & 0.3159  & 3.67  & 29.16 & 0.73  & 21.59  \\
    Janus-Pro-7B+GRPO & 0.7478 &	0.3999 &	0.5849 &	0.2481 &	0.3090 & 	0.3744 & 3.70 & 30.64 & 0.99  & \colorbox{mycolor_green}{22.12}   \\
    \hline
    \multicolumn{11}{c}{\textit{Our}} \\
    \hline
    LlamaGen+GCPO & 0.4691 &	0.4351 & 	0.5726 & 	0.1015 & 	0.3086 &	0.3199  & 2.97 & 26.27 & 0.10  & 20.59  \\            
    Janus-Pro-1B+GCPO & 0.7373 &	0.3201 &	0.4803 &	0.3220 &	0.2948 &	0.3059 & \colorbox{mycolor_green}{3.73} & 29.61 & 0.73 & 21.60  \\
    Janus-Pro-7B+GCPO & \colorbox{mycolor_green}{0.7508} &	\colorbox{mycolor_green}{0.5173} & 	\colorbox{mycolor_green}{0.7030} &	\colorbox{mycolor_green}{0.3824} &	\colorbox{mycolor_green}{0.3133} 	& \colorbox{mycolor_green}{0.3888} & \colorbox{mycolor_green}{3.73} & \colorbox{mycolor_green}{30.90}  & \colorbox{mycolor_green}{1.01}  & 22.10  \\
    \bottomrule
\end{tabular}
}}
\end{table}


    



\subsection{Main Results}
We compare our method with leading AR models and diffusion models on GenEval, as shown in Table~\ref{tab:generation_geneval}. Only utilizing critical tokens (30\% of the total tokens), GCPO achieves significant improvements over GRPO across all three base models. Notably, Janus-Pro-7B+GCPO attained the highest overall score of \textbf{0.90}. This is primarily attributed to a substantial improvement in the Counting task (+0.19). Furthermore, Table~\ref{tab:t2icompbench_quality} shows results of our method on T2I-CompBench, which differs substantially from the GenEval-style training data. GCPO consistently outperforms GRPO in the majority of tasks, achieving up to 20\% performance gains on Shape (+0.1174), Texture (+0.1181), and Spatial task (+0.1343), thereby demonstrating strong generalization.  As shown in Fig.~\ref{fig:visual_compare}, the model with our method accurately understands the shape of the mountain (big) and the tree (small). In contrast, the model with GRPO generates a forest.

As shown in Table~\ref{tab:t2icompbench_quality}, under the HPS-based reward setting, our method consistently outperforms GRPO and demonstrates superior performance across all three models. Notably, Janus-Pro-7B+GCPO achieves the best scores on image quality and human preference alignment. Meanwhile, the images generated by our method exhibit more natural and vivid details, as illustrated in Fig.~\ref{fig:visual_compare}. For more comparison results, please refer to Appendix~\ref{sec:More Visual Comparison Result}.

\begin{figure}[htbp]
    \centering
    \subcaptionbox{Comparison on token selection strategies and selection ratios.\label{fig:select_ratio:a}}[.32\linewidth]{
        \includegraphics[width=\linewidth]{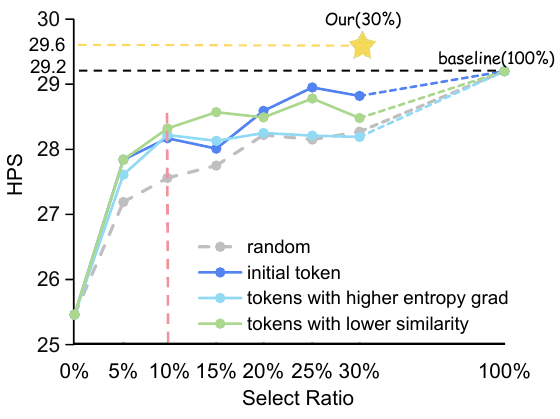}
    }
    \hfill
    \subcaptionbox{Comparison of critical tokens and other tokens on HPS\label{fig:select_ratio:b}}[.32\linewidth]{
        \includegraphics[width=\linewidth]{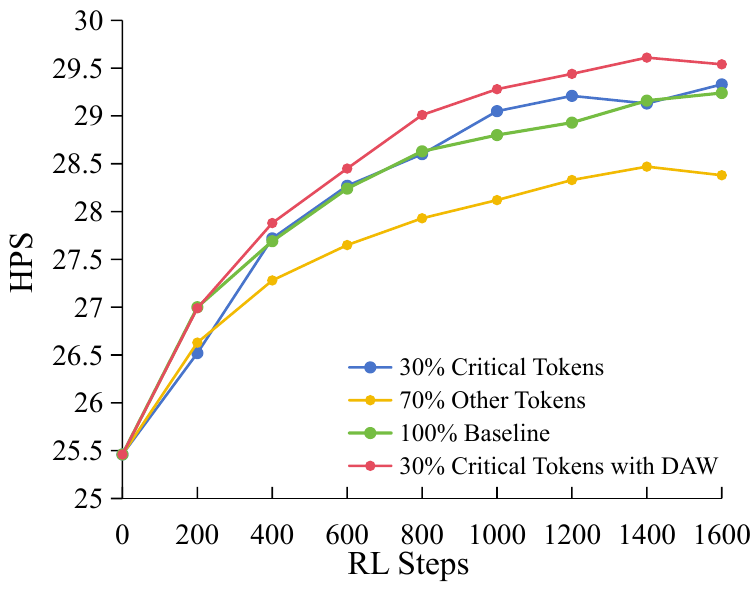}
    }
    \hfill
    \subcaptionbox{Comparison of critical tokens and other tokens on Geneval\label{fig:select_ratio:c}}[.32\linewidth]{
        \includegraphics[width=\linewidth]{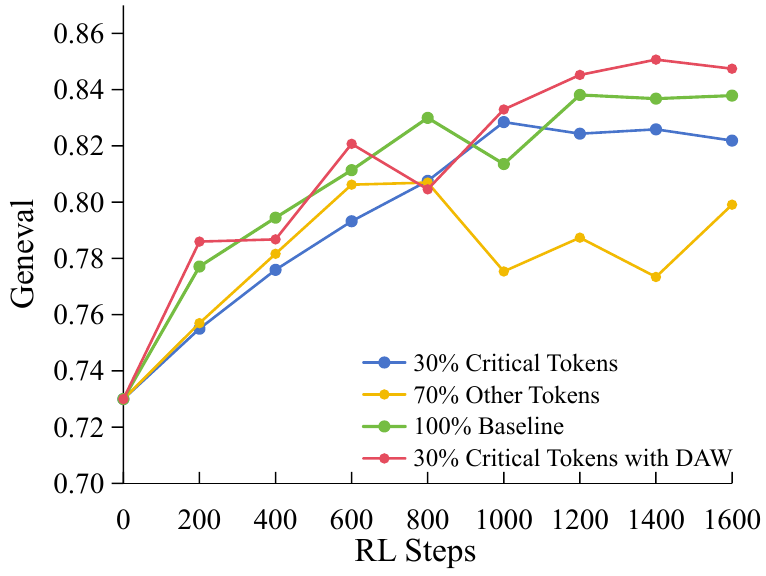}
    }
    \caption{(a): All three types of critical tokens deliver clear performance gains at 10\% initial selection ratios and outperform random selection, with further increases yield only limited improvements. (b) \& (c): GRPO with critical tokens (30\%) has more performance improvement than GRPO with remaining tokens (70\%). Our GCPO further achieves performance improvement.}
    \label{fig:select_ratio}
\end{figure}

\subsection{Ablation Study}
\label{Ablation Study}
We conduct analysis study on different token-selection strategies and selection rates, as shown in Fig.~\ref{fig:select_ratio:a}. Within the initial selection ratios range (10\% of tokens), GRPO performance rises rapidly (+2.81) with each of the three types of critical tokens, respectively; beyond this threshold, the gains are limited (+0.95). Moreover, combining all three types of critical tokens clearly outperforms selecting any single type at the same token budget (30\% of tokens), which demonstrates the correct identification of critical tokens in RLVL training. We further verify that all three strategies significantly outperform random selection. As shown in Table~\ref{tab:ablation}, we present more experimental results on multiple benchmarks under Geneval and HPS reward settings. Our critical token strategies consistently achieve balanced performance improvements. Furthermore, with the introduction of dynamic advantage weight, all metrics reach optimal values, demonstrating the effectiveness of our designs.

We also consider the comparison between 30\% critical tokens and 70\% of the remaining tokens. As shown in Fig.~\ref{fig:select_ratio:b} and~\ref{fig:select_ratio:c}, GRPO with critical tokens are comparable with the GRPO baseline, while GRPO with other tokens leads to a drop in performance. Despite the fact that the other tokens account for 70\% of the total tokens used in training and more than twice the count of critical tokens, they still have a significant performance gap. This finding highlights the greater importance of critical tokens based on our selection strategy for effective model training.

\begin{table}[h]
\centering
\caption{Ablation results on critical tokens and dynamic advantage weight. Init-T: initial tokens, HG-T: high entropy gradient tokens, LS-T: low similarity tokens; DAW: dynamic advantage weight.}
\label{tab:ablation}

\small
\setlength{\tabcolsep}{1.5mm}{
\resizebox{0.99\linewidth}{!}{
\begin{tabular}{cccccccccccccc}
\toprule
\addlinespace[0.3ex]
\multirow{2}{*}{Init-T} & \multirow{2}{*}{HG-T} & \multirow{2}{*}{LS-T} & \multirow{2}{*}{DAW} & \multirow{2}{*}{GenEval$\uparrow$} & \multicolumn{6}{c}{\rule{0pt}{2.4ex}T2I-CompBench} & \multirow{2}{*}{DEQA$\uparrow$} & \multirow{2}{*}{HPS$\uparrow$} & \multirow{2}{*}{ImgRwd$\uparrow$} \\
\cmidrule(lr){6-11}
& & & & & Color$\uparrow$ & Shape$\uparrow$ & Texture$\uparrow$ & Spatial$\uparrow$ & Non-Spat.$\uparrow$ & Complex$\uparrow$ & & & \\
\midrule
$\checkmark$ & - & - & -  & 0.82 & 0.6084 & 0.2820 & 0.3496 & 0.2374  & 0.2865 & 0.2877 & 3.66 & 28.90 & 0.63 \\
- & $\checkmark$ & - & -  & 0.81 & 0.5440 & 0.2712 & 0.3368 & 0.2262 & 0.2832 & 0.2794 & 3.63 & 28.22 & 0.63 \\
- & - & $\checkmark$ & -  & 0.82 & 0.6721 & 0.2942 & 0.4101 & 0.2569 & 0.2938 & 0.3109 & 3.64  & 28.78  & 0.66 \\
$\checkmark$ & $\checkmark$ & $\checkmark$ & -  & 0.83 & 0.6602 & 0.2924  & 0.3991  & 0.2936  & 0.2943 & 0.3015 & 3.70  & 29.33 & 0.71 \\
$\checkmark$ & $\checkmark$ & $\checkmark$ & $\checkmark$  & 0.85 & 0.7373 & 0.3201 & 0.4803 & 0.3220 & 0.2948 & 0.3059 & 3.73  & 29.61 & 0.73 \\
\bottomrule
\end{tabular}
}}
\end{table}

\section{Conclusion}
\label{sec:Conclusion}
In this paper, we introduce Group Critical-token Policy Optimization (\textbf{GCPO}), a novel RLVR framework for autoregressive image generation. We identify critical tokens in RLVR-based AR visual generation from three perspectives: Causal dependency of AR, Entropy-induced spatial structure, and RLVR-focused token diversity. We devise the dynamic advantage weight for critical tokens to enable reasonable exploration, based on their confidence divergence between the policy model and reference model. Extensive experiments demonstrate that by leveraging critical tokens (30\% of the image tokens), GCPO achieves better performance than GRPO, which operates on the full tokens. 

\clearpage




\bibliography{iclr2026_conference}

\begin{thebibliography}{58}
\providecommand{\natexlab}[1]{#1}
\providecommand{\url}[1]{\texttt{#1}}
\expandafter\ifx\csname urlstyle\endcsname\relax
  \providecommand{\doi}[1]{doi: #1}\else
  \providecommand{\doi}{doi: \begingroup \urlstyle{rm}\Url}\fi

\bibitem[Betker et~al.(2023)Betker, Goh, Jing, Brooks, Wang, Li, Ouyang,
  Zhuang, Lee, Guo, et~al.]{betker2023improving}
James Betker, Gabriel Goh, Li~Jing, Tim Brooks, Jianfeng Wang, Linjie Li, Long
  Ouyang, Juntang Zhuang, Joyce Lee, Yufei Guo, et~al.
\newblock Improving image generation with better captions.
\newblock \emph{Computer Science. https://cdn. openai. com/papers/dall-e-3.
  pdf}, 2\penalty0 (3):\penalty0 8, 2023.

\bibitem[Beyer et~al.(2025)Beyer, Li, Chen, Karaman, and He]{beyer2025highly}
L~Lao Beyer, Tianhong Li, Xinlei Chen, Sertac Karaman, and Kaiming He.
\newblock Highly compressed tokenizer can generate without training.
\newblock \emph{arXiv preprint arXiv:2506.08257}, 2025.

\bibitem[briaai(2025)]{RMBG}
briaai.
\newblock Rmbg v2.0 is a new state-of-the-art background removal model.
\newblock \url{https://huggingface.co/briaai/RMBG-2.0/}, 2025.

\bibitem[Chen et~al.(2024)Chen, Wu, Luo, Xie, Paul, Luo, Zhao, and
  Li]{chen2024pixart}
Junsong Chen, Yue Wu, Simian Luo, Enze Xie, Sayak Paul, Ping Luo, Hang Zhao,
  and Zhenguo Li.
\newblock Pixart-$\{$$\backslash$delta$\}$: Fast and controllable image
  generation with latent consistency models.
\newblock \emph{arXiv preprint arXiv:2401.05252}, 2024.

\bibitem[Chen et~al.(2025{\natexlab{a}})Chen, Wu, Liu, Pan, Liu, Xie, Yu, and
  Ruan]{chen2025janus}
Xiaokang Chen, Zhiyu Wu, Xingchao Liu, Zizheng Pan, Wen Liu, Zhenda Xie,
  Xingkai Yu, and Chong Ruan.
\newblock Janus-pro: Unified multimodal understanding and generation with data
  and model scaling.
\newblock \emph{arXiv preprint arXiv:2501.17811}, 2025{\natexlab{a}}.

\bibitem[Chen et~al.(2025{\natexlab{b}})Chen, Zhu, Qiu, Dong, Wang, Wu, Li,
  Sotiras, Wang, and Razi]{chen2025dra}
Xiwen Chen, Wenhui Zhu, Peijie Qiu, Xuanzhao Dong, Hao Wang, Haiyu Wu, Huayu
  Li, Aristeidis Sotiras, Yalin Wang, and Abolfazl Razi.
\newblock Dra-grpo: Exploring diversity-aware reward adjustment for
  r1-zero-like training of large language models.
\newblock \emph{arXiv preprint arXiv:2505.09655}, 2025{\natexlab{b}}.

\bibitem[Deng et~al.(2025)Deng, Zhu, Li, Gou, Li, Wang, Zhong, Yu, Nie, Song,
  et~al.]{deng2025emerging}
Chaorui Deng, Deyao Zhu, Kunchang Li, Chenhui Gou, Feng Li, Zeyu Wang, Shu
  Zhong, Weihao Yu, Xiaonan Nie, Ziang Song, et~al.
\newblock Emerging properties in unified multimodal pretraining.
\newblock \emph{arXiv preprint arXiv:2505.14683}, 2025.

\bibitem[Esser et~al.(2021)Esser, Rombach, and Ommer]{esser2021taming}
Patrick Esser, Robin Rombach, and Bjorn Ommer.
\newblock Taming transformers for high-resolution image synthesis.
\newblock In \emph{Proceedings of the IEEE/CVF conference on computer vision
  and pattern recognition}, pp.\  12873--12883, 2021.

\bibitem[Esser et~al.(2024)Esser, Kulal, Blattmann, Entezari, M{\"u}ller,
  Saini, Levi, Lorenz, Sauer, Boesel, et~al.]{esser2024scaling}
Patrick Esser, Sumith Kulal, Andreas Blattmann, Rahim Entezari, Jonas
  M{\"u}ller, Harry Saini, Yam Levi, Dominik Lorenz, Axel Sauer, Frederic
  Boesel, et~al.
\newblock Scaling rectified flow transformers for high-resolution image
  synthesis.
\newblock In \emph{Forty-first international conference on machine learning},
  2024.

\bibitem[Ghosh et~al.(2023)Ghosh, Hajishirzi, and Schmidt]{ghosh2023geneval}
Dhruba Ghosh, Hannaneh Hajishirzi, and Ludwig Schmidt.
\newblock Geneval: An object-focused framework for evaluating text-to-image
  alignment.
\newblock \emph{Advances in Neural Information Processing Systems},
  36:\penalty0 52132--52152, 2023.

\bibitem[Guo et~al.(2025{\natexlab{a}})Guo, Yang, Zhang, Song, Zhang, Xu, Zhu,
  Ma, Wang, Bi, et~al.]{guo2025deepseek}
Daya Guo, Dejian Yang, Haowei Zhang, Junxiao Song, Ruoyu Zhang, Runxin Xu,
  Qihao Zhu, Shirong Ma, Peiyi Wang, Xiao Bi, et~al.
\newblock Deepseek-r1: Incentivizing reasoning capability in llms via
  reinforcement learning.
\newblock \emph{arXiv preprint arXiv:2501.12948}, 2025{\natexlab{a}}.

\bibitem[Guo et~al.(2025{\natexlab{b}})Guo, Zhang, Tong, Zhao, Huang, Zhang,
  Zhang, Liu, Zhang, Gao, et~al.]{guo2025can}
Ziyu Guo, Renrui Zhang, Chengzhuo Tong, Zhizheng Zhao, Rui Huang, Haoquan
  Zhang, Manyuan Zhang, Jiaming Liu, Shanghang Zhang, Peng Gao, et~al.
\newblock Can we generate images with cot? let's verify and reinforce image
  generation step by step.
\newblock \emph{arXiv preprint arXiv:2501.13926}, 2025{\natexlab{b}}.

\bibitem[He et~al.(2025)He, Fried, and Welleck]{he2025rewarding}
Andre He, Daniel Fried, and Sean Welleck.
\newblock Rewarding the unlikely: Lifting grpo beyond distribution sharpening.
\newblock \emph{arXiv preprint arXiv:2506.02355}, 2025.

\bibitem[He et~al.(2024)He, Chen, He, He, Zhou, Zhang, and Zhuang]{he2024zipar}
Yefei He, Feng Chen, Yuanyu He, Shaoxuan He, Hong Zhou, Kaipeng Zhang, and
  Bohan Zhuang.
\newblock Zipar: Accelerating autoregressive image generation through spatial
  locality.
\newblock \emph{arXiv preprint arXiv:2412.04062}, 2\penalty0 (3):\penalty0 4,
  2024.

\bibitem[Huang et~al.(2023)Huang, Sun, Xie, Li, and Liu]{huang2023t2i}
Kaiyi Huang, Kaiyue Sun, Enze Xie, Zhenguo Li, and Xihui Liu.
\newblock T2i-compbench: A comprehensive benchmark for open-world compositional
  text-to-image generation.
\newblock \emph{Advances in Neural Information Processing Systems},
  36:\penalty0 78723--78747, 2023.

\bibitem[Jiang et~al.(2025)Jiang, Guo, Zhang, Zong, Li, Zhuo, Yan, Heng, and
  Li]{jiang2025t2i}
Dongzhi Jiang, Ziyu Guo, Renrui Zhang, Zhuofan Zong, Hao Li, Le~Zhuo, Shilin
  Yan, Pheng-Ann Heng, and Hongsheng Li.
\newblock T2i-r1: Reinforcing image generation with collaborative
  semantic-level and token-level cot.
\newblock \emph{arXiv preprint arXiv:2505.00703}, 2025.

\bibitem[Kirstain et~al.(2023)Kirstain, Polyak, Singer, Matiana, Penna, and
  Levy]{kirstain2023pick}
Yuval Kirstain, Adam Polyak, Uriel Singer, Shahbuland Matiana, Joe Penna, and
  Omer Levy.
\newblock Pick-a-pic: An open dataset of user preferences for text-to-image
  generation.
\newblock \emph{Advances in neural information processing systems},
  36:\penalty0 36652--36663, 2023.

\bibitem[Labs(2024)]{Flux}
Black~Forest Labs.
\newblock Flux.
\newblock \url{https://github.com/black-forest-labs/flux/}, 2024.

\bibitem[Li et~al.(2024)Li, Zhang, Guo, Zhang, Li, Zhang, Zhang, Zhang, Li,
  Liu, et~al.]{li2024llava}
Bo~Li, Yuanhan Zhang, Dong Guo, Renrui Zhang, Feng Li, Hao Zhang, Kaichen
  Zhang, Peiyuan Zhang, Yanwei Li, Ziwei Liu, et~al.
\newblock Llava-onevision: Easy visual task transfer.
\newblock \emph{arXiv preprint arXiv:2408.03326}, 2024.

\bibitem[Li et~al.(2025)Li, Cao, Griggs, Liu, Mo, Tang, Hegde, Hakhamaneshi,
  Patil, Zaharia, et~al.]{li2025llms}
Dacheng Li, Shiyi Cao, Tyler Griggs, Shu Liu, Xiangxi Mo, Eric Tang, Sumanth
  Hegde, Kourosh Hakhamaneshi, Shishir~G Patil, Matei Zaharia, et~al.
\newblock Llms can easily learn to reason from demonstrations structure, not
  content, is what matters!
\newblock \emph{arXiv preprint arXiv:2502.07374}, 2025.

\bibitem[Lin et~al.(2024)Lin, Liang, Xu, Lin, Wang, Luo, Shi, Li, Yang, and
  Tu]{lin2024critical}
Zicheng Lin, Tian Liang, Jiahao Xu, Qiuzhi Lin, Xing Wang, Ruilin Luo, Chufan
  Shi, Siheng Li, Yujiu Yang, and Zhaopeng Tu.
\newblock Critical tokens matter: Token-level contrastive estimation enhances
  llm's reasoning capability.
\newblock \emph{arXiv preprint arXiv:2411.19943}, 2024.

\bibitem[Liu et~al.(2024{\natexlab{a}})Liu, Zhao, Zhuo, Lin, Xin, Li, Qin,
  Qiao, Li, and Gao]{liu2024lumina}
Dongyang Liu, Shitian Zhao, Le~Zhuo, Weifeng Lin, Yi~Xin, Xinyue Li, Qi~Qin,
  Yu~Qiao, Hongsheng Li, and Peng Gao.
\newblock Lumina-mgpt: Illuminate flexible photorealistic text-to-image
  generation with multimodal generative pretraining.
\newblock \emph{arXiv preprint arXiv:2408.02657}, 2024{\natexlab{a}}.

\bibitem[Liu et~al.(2025)Liu, Liu, Liang, Li, Liu, Wang, Wan, Zhang, and
  Ouyang]{liu2025flow}
Jie Liu, Gongye Liu, Jiajun Liang, Yangguang Li, Jiaheng Liu, Xintao Wang,
  Pengfei Wan, Di~Zhang, and Wanli Ouyang.
\newblock Flow-grpo: Training flow matching models via online rl.
\newblock \emph{arXiv preprint arXiv:2505.05470}, 2025.

\bibitem[Liu et~al.(2024{\natexlab{b}})Liu, Zeng, Ren, Li, Zhang, Yang, Jiang,
  Li, Yang, Su, et~al.]{liu2024grounding}
Shilong Liu, Zhaoyang Zeng, Tianhe Ren, Feng Li, Hao Zhang, Jie Yang, Qing
  Jiang, Chunyuan Li, Jianwei Yang, Hang Su, et~al.
\newblock Grounding dino: Marrying dino with grounded pre-training for open-set
  object detection.
\newblock In \emph{European conference on computer vision}, pp.\  38--55.
  Springer, 2024{\natexlab{b}}.

\bibitem[Ma et~al.(2024)Ma, Zhou, Liang, Bai, Zhao, Li, Chen, and
  Jin]{ma2024star}
Xiaoxiao Ma, Mohan Zhou, Tao Liang, Yalong Bai, Tiejun Zhao, Biye Li, Huaian
  Chen, and Yi~Jin.
\newblock Star: Scale-wise text-conditioned autoregressive image generation.
\newblock \emph{arXiv preprint arXiv:2406.10797}, 2024.

\bibitem[OpenAI(2024)]{openaio1}
OpenAI.
\newblock Learning to reason with llms.
\newblock \url{https://openai.com/index/learning-to-reason-with-llms/}, 2024.

\bibitem[OpenAI(2025)]{chatgpt4o}
OpenAI.
\newblock Introducing 4o image generation.
\newblock \url{https://openai.com/index/introducing-4o-image-generation/},
  2025.

\bibitem[Pan et~al.(2025)Pan, Bu, Wu, Wu, Shen, Li, Zhao, Li, Tang, and
  Zhuang]{pan2025focusdiff}
Kaihang Pan, Wendong Bu, Yuruo Wu, Yang Wu, Kai Shen, Yunfei Li, Hang Zhao,
  Juncheng Li, Siliang Tang, and Yueting Zhuang.
\newblock Focusdiff: Advancing fine-grained text-image alignment for
  autoregressive visual generation through rl.
\newblock \emph{arXiv preprint arXiv:2506.05501}, 2025.

\bibitem[Podell et~al.(2023)Podell, English, Lacey, Blattmann, Dockhorn,
  M{\"u}ller, Penna, and Rombach]{podell2023sdxl}
Dustin Podell, Zion English, Kyle Lacey, Andreas Blattmann, Tim Dockhorn, Jonas
  M{\"u}ller, Joe Penna, and Robin Rombach.
\newblock Sdxl: Improving latent diffusion models for high-resolution image
  synthesis.
\newblock \emph{arXiv preprint arXiv:2307.01952}, 2023.

\bibitem[Ramesh et~al.(2022)Ramesh, Dhariwal, Nichol, Chu, and
  Chen]{ramesh2022hierarchical}
Aditya Ramesh, Prafulla Dhariwal, Alex Nichol, Casey Chu, and Mark Chen.
\newblock Hierarchical text-conditional image generation with clip latents.
\newblock \emph{arXiv preprint arXiv:2204.06125}, 1\penalty0 (2):\penalty0 3,
  2022.

\bibitem[Saharia et~al.(2022)Saharia, Chan, Saxena, Li, Whang, Denton,
  Ghasemipour, Gontijo~Lopes, Karagol~Ayan, Salimans,
  et~al.]{saharia2022photorealistic}
Chitwan Saharia, William Chan, Saurabh Saxena, Lala Li, Jay Whang, Emily~L
  Denton, Kamyar Ghasemipour, Raphael Gontijo~Lopes, Burcu Karagol~Ayan, Tim
  Salimans, et~al.
\newblock Photorealistic text-to-image diffusion models with deep language
  understanding.
\newblock \emph{Advances in neural information processing systems},
  35:\penalty0 36479--36494, 2022.

\bibitem[Shannon(1948)]{shannon1948mathematical}
Claude~E Shannon.
\newblock A mathematical theory of communication.
\newblock \emph{The Bell system technical journal}, 27\penalty0 (3):\penalty0
  379--423, 1948.

\bibitem[Shao et~al.(2024)Shao, Wang, Zhu, Xu, Song, Bi, Zhang, Zhang, Li, Wu,
  et~al.]{shao2024deepseekmath}
Zhihong Shao, Peiyi Wang, Qihao Zhu, Runxin Xu, Junxiao Song, Xiao Bi, Haowei
  Zhang, Mingchuan Zhang, YK~Li, Yang Wu, et~al.
\newblock Deepseekmath: Pushing the limits of mathematical reasoning in open
  language models.
\newblock \emph{arXiv preprint arXiv:2402.03300}, 2024.

\bibitem[Shrivastava et~al.(2025)Shrivastava, Awadallah, Balachandran, Garg,
  Behl, and Papailiopoulos]{shrivastava2025sample}
Vaishnavi Shrivastava, Ahmed Awadallah, Vidhisha Balachandran, Shivam Garg,
  Harkirat Behl, and Dimitris Papailiopoulos.
\newblock Sample more to think less: Group filtered policy optimization for
  concise reasoning.
\newblock \emph{arXiv preprint arXiv:2508.09726}, 2025.

\bibitem[Sun et~al.(2024)Sun, Jiang, Chen, Zhang, Peng, Luo, and
  Yuan]{sun2024autoregressive}
Peize Sun, Yi~Jiang, Shoufa Chen, Shilong Zhang, Bingyue Peng, Ping Luo, and
  Zehuan Yuan.
\newblock Autoregressive model beats diffusion: Llama for scalable image
  generation.
\newblock \emph{arXiv preprint arXiv:2406.06525}, 2024.

\bibitem[Team(2024)]{team2024chameleon}
Chameleon Team.
\newblock Chameleon: Mixed-modal early-fusion foundation models.
\newblock \emph{arXiv preprint arXiv:2405.09818}, 2024.

\bibitem[Team et~al.(2025)Team, Du, Gao, Xing, Jiang, Chen, Li, Xiao, Du, Liao,
  et~al.]{team2025kimi}
Kimi Team, Angang Du, Bofei Gao, Bowei Xing, Changjiu Jiang, Cheng Chen, Cheng
  Li, Chenjun Xiao, Chenzhuang Du, Chonghua Liao, et~al.
\newblock Kimi k1. 5: Scaling reinforcement learning with llms.
\newblock \emph{arXiv preprint arXiv:2501.12599}, 2025.

\bibitem[Van Den~Oord et~al.(2017)Van Den~Oord, Vinyals, et~al.]{van2017neural}
Aaron Van Den~Oord, Oriol Vinyals, et~al.
\newblock Neural discrete representation learning.
\newblock \emph{Advances in neural information processing systems}, 30, 2017.

\bibitem[Vassoyan et~al.(2025)Vassoyan, Beau, and Plaud]{vassoyan2025ignore}
Jean Vassoyan, Nathana{\"e}l Beau, and Roman Plaud.
\newblock Ignore the kl penalty! boosting exploration on critical tokens to
  enhance rl fine-tuning.
\newblock \emph{arXiv preprint arXiv:2502.06533}, 2025.

\bibitem[Wang et~al.(2025{\natexlab{a}})Wang, Liu, Zhang, Li, and
  Zhou]{wang2025stabilizing}
Jiakang Wang, Runze Liu, Fuzheng Zhang, Xiu Li, and Guorui Zhou.
\newblock Stabilizing knowledge, promoting reasoning: Dual-token constraints
  for rlvr.
\newblock \emph{arXiv preprint arXiv:2507.15778}, 2025{\natexlab{a}}.

\bibitem[Wang et~al.(2022)Wang, Yang, Hu, Li, Lin, Gan, Liu, Liu, and
  Wang]{wang2022git}
Jianfeng Wang, Zhengyuan Yang, Xiaowei Hu, Linjie Li, Kevin Lin, Zhe Gan,
  Zicheng Liu, Ce~Liu, and Lijuan Wang.
\newblock Git: A generative image-to-text transformer for vision and language.
\newblock \emph{arXiv preprint arXiv:2205.14100}, 2022.

\bibitem[Wang et~al.(2025{\natexlab{b}})Wang, Tian, Wang, Zhang, Huang, Wu, and
  Jiang]{wang2025simplear}
Junke Wang, Zhi Tian, Xun Wang, Xinyu Zhang, Weilin Huang, Zuxuan Wu, and
  Yu-Gang Jiang.
\newblock Simplear: Pushing the frontier of autoregressive visual generation
  through pretraining, sft, and rl.
\newblock \emph{arXiv preprint arXiv:2504.11455}, 2025{\natexlab{b}}.

\bibitem[Wang et~al.(2025{\natexlab{c}})Wang, Yu, Gao, Zheng, Liu, Lu, Dang,
  Chen, Yang, Zhang, et~al.]{wang2025beyond}
Shenzhi Wang, Le~Yu, Chang Gao, Chujie Zheng, Shixuan Liu, Rui Lu, Kai Dang,
  Xionghui Chen, Jianxin Yang, Zhenru Zhang, et~al.
\newblock Beyond the 80/20 rule: High-entropy minority tokens drive effective
  reinforcement learning for llm reasoning.
\newblock \emph{arXiv preprint arXiv:2506.01939}, 2025{\natexlab{c}}.

\bibitem[Wang et~al.(2024)Wang, Zhang, Luo, Sun, Cui, Wang, Zhang, Wang, Li,
  Yu, et~al.]{wang2024emu3}
Xinlong Wang, Xiaosong Zhang, Zhengxiong Luo, Quan Sun, Yufeng Cui, Jinsheng
  Wang, Fan Zhang, Yueze Wang, Zhen Li, Qiying Yu, et~al.
\newblock Emu3: Next-token prediction is all you need.
\newblock \emph{arXiv preprint arXiv:2409.18869}, 2024.

\bibitem[Wu et~al.(2023)Wu, Hao, Sun, Chen, Zhu, Zhao, and Li]{wu2023human}
Xiaoshi Wu, Yiming Hao, Keqiang Sun, Yixiong Chen, Feng Zhu, Rui Zhao, and
  Hongsheng Li.
\newblock Human preference score v2: A solid benchmark for evaluating human
  preferences of text-to-image synthesis.
\newblock \emph{arXiv preprint arXiv:2306.09341}, 2023.

\bibitem[Xiang \& Fan(2025)Xiang and Fan]{xiang2025make}
Xunzhi Xiang and Qi~Fan.
\newblock Make it efficient: Dynamic sparse attention for autoregressive image
  generation.
\newblock \emph{arXiv preprint arXiv:2506.18226}, 2025.

\bibitem[Xie et~al.(2024)Xie, Mao, Bai, Zhang, Wang, Lin, Gu, Chen, Yang, and
  Shou]{xie2024show}
Jinheng Xie, Weijia Mao, Zechen Bai, David~Junhao Zhang, Weihao Wang,
  Kevin~Qinghong Lin, Yuchao Gu, Zhijie Chen, Zhenheng Yang, and Mike~Zheng
  Shou.
\newblock Show-o: One single transformer to unify multimodal understanding and
  generation.
\newblock \emph{arXiv preprint arXiv:2408.12528}, 2024.

\bibitem[Xu et~al.(2023)Xu, Liu, Wu, Tong, Li, Ding, Tang, and
  Dong]{xu2023imagereward}
Jiazheng Xu, Xiao Liu, Yuchen Wu, Yuxuan Tong, Qinkai Li, Ming Ding, Jie Tang,
  and Yuxiao Dong.
\newblock Imagereward: Learning and evaluating human preferences for
  text-to-image generation.
\newblock \emph{Advances in Neural Information Processing Systems},
  36:\penalty0 15903--15935, 2023.

\bibitem[Xue et~al.(2025)Xue, Wu, Gao, Kong, Zhu, Chen, Liu, Liu, Guo, Huang,
  et~al.]{xue2025dancegrpo}
Zeyue Xue, Jie Wu, Yu~Gao, Fangyuan Kong, Lingting Zhu, Mengzhao Chen, Zhiheng
  Liu, Wei Liu, Qiushan Guo, Weilin Huang, et~al.
\newblock Dancegrpo: Unleashing grpo on visual generation.
\newblock \emph{arXiv preprint arXiv:2505.07818}, 2025.

\bibitem[Yang et~al.(2025)Yang, Li, Yang, Zhang, Hui, Zheng, Yu, Gao, Huang,
  Lv, et~al.]{yang2025qwen3}
An~Yang, Anfeng Li, Baosong Yang, Beichen Zhang, Binyuan Hui, Bo~Zheng, Bowen
  Yu, Chang Gao, Chengen Huang, Chenxu Lv, et~al.
\newblock Qwen3 technical report.
\newblock \emph{arXiv preprint arXiv:2505.09388}, 2025.

\bibitem[Yoon et~al.(2025)Yoon, Yoon, Hasegawa-Johnson, Kim, and
  Yoo]{yoonconfpo}
Hee~Suk Yoon, Eunseop Yoon, Mark~A Hasegawa-Johnson, Sungwoong Kim, and Chang~D
  Yoo.
\newblock Confpo: Exploiting policy model confidence for critical token
  selection in preference optimization.
\newblock In \emph{Forty-second International Conference on Machine Learning},
  2025.

\bibitem[You et~al.(2025)You, Cai, Gu, Xue, and Dong]{you2025teaching}
Zhiyuan You, Xin Cai, Jinjin Gu, Tianfan Xue, and Chao Dong.
\newblock Teaching large language models to regress accurate image quality
  scores using score distribution.
\newblock In \emph{Proceedings of the Computer Vision and Pattern Recognition
  Conference}, pp.\  14483--14494, 2025.

\bibitem[Yu et~al.(2021)Yu, Li, Koh, Zhang, Pang, Qin, Ku, Xu, Baldridge, and
  Wu]{yu2021vector}
Jiahui Yu, Xin Li, Jing~Yu Koh, Han Zhang, Ruoming Pang, James Qin, Alexander
  Ku, Yuanzhong Xu, Jason Baldridge, and Yonghui Wu.
\newblock Vector-quantized image modeling with improved vqgan.
\newblock \emph{arXiv preprint arXiv:2110.04627}, 2021.

\bibitem[Yu et~al.(2025)Yu, Zhang, Zhu, Yuan, Zuo, Yue, Dai, Fan, Liu, Liu,
  et~al.]{yu2025dapo}
Qiying Yu, Zheng Zhang, Ruofei Zhu, Yufeng Yuan, Xiaochen Zuo, Yu~Yue, Weinan
  Dai, Tiantian Fan, Gaohong Liu, Lingjun Liu, et~al.
\newblock Dapo: An open-source llm reinforcement learning system at scale.
\newblock \emph{arXiv preprint arXiv:2503.14476}, 2025.

\bibitem[Yuan et~al.(2025)Yuan, Liu, Yue, Zhang, Zuo, Wang, Zhang, and
  Zhou]{yuan2025ar}
Shihao Yuan, Yahui Liu, Yang Yue, Jingyuan Zhang, Wangmeng Zuo, Qi~Wang,
  Fuzheng Zhang, and Guorui Zhou.
\newblock Ar-grpo: Training autoregressive image generation models via
  reinforcement learning.
\newblock \emph{arXiv preprint arXiv:2508.06924}, 2025.

\bibitem[Yue et~al.(2025)Yue, Yuan, Yu, Zuo, Zhu, Xu, Chen, Wang, Fan, Du,
  et~al.]{yue2025vapo}
Yu~Yue, Yufeng Yuan, Qiying Yu, Xiaochen Zuo, Ruofei Zhu, Wenyuan Xu, Jiaze
  Chen, Chengyi Wang, TianTian Fan, Zhengyin Du, et~al.
\newblock Vapo: Efficient and reliable reinforcement learning for advanced
  reasoning tasks.
\newblock \emph{arXiv preprint arXiv:2504.05118}, 2025.

\bibitem[Zhang et~al.(2025)Zhang, Li, Yang, Wang, Yang, Qi, Bao, Chen, Luo, and
  Qiu]{zhang2025reasongen}
Yu~Zhang, Yunqi Li, Yifan Yang, Rui Wang, Yuqing Yang, Dai Qi, Jianmin Bao,
  Dongdong Chen, Chong Luo, and Lili Qiu.
\newblock Reasongen-r1: Cot for autoregressive image generation models through
  sft and rl.
\newblock \emph{arXiv preprint arXiv:2505.24875}, 2025.

\bibitem[Zhou et~al.(2024)Zhou, Yu, Babu, Tirumala, Yasunaga, Shamis, Kahn, Ma,
  Zettlemoyer, and Levy]{zhou2024transfusion}
Chunting Zhou, Lili Yu, Arun Babu, Kushal Tirumala, Michihiro Yasunaga, Leonid
  Shamis, Jacob Kahn, Xuezhe Ma, Luke Zettlemoyer, and Omer Levy.
\newblock Transfusion: Predict the next token and diffuse images with one
  multi-modal model.
\newblock \emph{arXiv preprint arXiv:2408.11039}, 2024.

\end{thebibliography}
\bibliographystyle{iclr2026_conference}

\clearpage

\appendix
\section{Detailed Experimental Setup}
\label{sec:Detailed Experimental Setup}

\subsection{HyperParameter}
In this paper, we used three models (LlamaGen~\citep{sun2024autoregressive}, Janus-Pro 1B~\citep{chen2025janus}, and Janus-Pro 7B~\citep{chen2025janus}) on the composition image task and image quality task. For different models and reward (HPS Reward~\citep{wu2023human} and Geneval Reward~\citep{liu2025flow})settings, our training configuration and parameters are as follows: 
\begin{table}[h]
\centering
\caption{GCPO training hyperparameters.}
\resizebox{0.99\linewidth}{!}{
\begin{tabular}{l|ccccc}
\toprule
\textbf{Name}  & \textbf{LlamaGen for HPS} &\textbf{1B for Geneval} & \textbf{1B for HPS} & \textbf{7B for Geneval} & \textbf{7B for HPS}\\
\midrule
Learning rate &1e-5 &3e-6  & 1e-6 & 3e-6  & 1e-6\\
Beta $\beta$ & 0.04 &0.01  & 0.01 & 0.01  & 0.01\\
Group Size $G$ & 8 &4 & 4 & 4 & 8\\
Classifier-Free Guidance Scale & 1 &5  & 5 & 5  & 5\\
Max Gradient Norm & 1.0 &1.0  & 1.0 & 1.0  & 1.0\\
Batchsize & 2 &2  & 2 & 2  & 4\\
Training Steps & 900 &1,600  & 1,600 & 1,600  & 1,600\\
Gradient Accumulation Steps & 2 &1 & 1 & 1 & 1\\
Dynamic advantage weight clip $\epsilon_{w}$  & 0.5 &1 & 0.5 & 0.6 & 1\\
Image Resolution $h \times w$ & $256 \times 256$ & $384 \times 384$ & $384 \times 384$& $384 \times 384$ & $384 \times 384$ \\
\bottomrule
\end{tabular}
}
\end{table}

\subsection{Training efficiency}
Compared with standard GRPO, our additional operations involve entropy gradient calculation and token embedding similarity computation. We have implemented certain optimizations, so these operations do not introduce additional computational burden or affect training efficiency.

\section{Entropy analysis in RLVR-based AR visual generation}
\label{sec:appendix_entropy_analysis}
\begin{figure}[h]
    \centering
    \subcaptionbox{The entropy distribution of simple or compositional prompts. The entropy of the background is significantly greater than that of the subject. \label{fig:entropy_low:a}}[.49\linewidth]{
        \includegraphics[width=\linewidth]{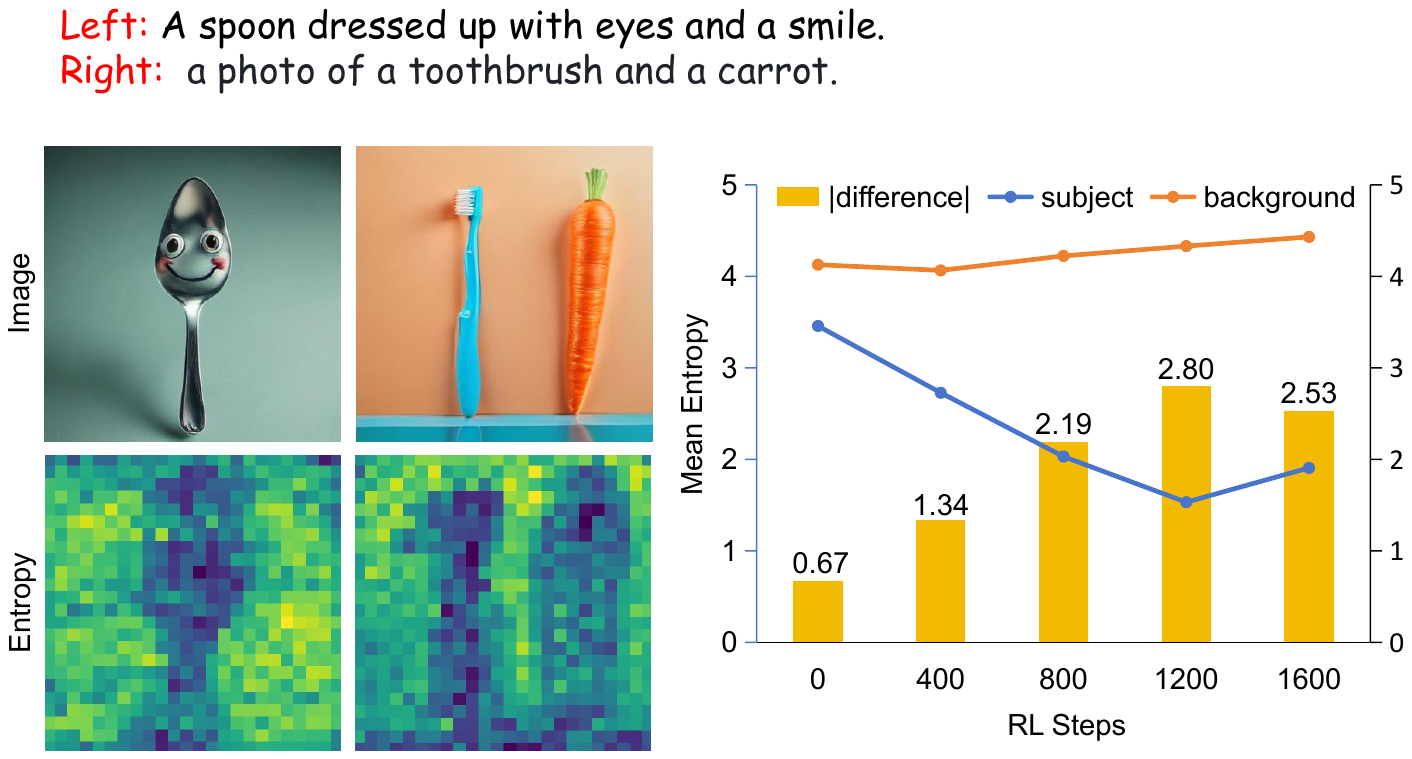}
    }
    \hfill
    \subcaptionbox{The entropy distribution of complex or detailed prompts. The entropy of the subject is greater than that of the background.\label{fig:entropy_low:b}}[.49\linewidth]{
        \includegraphics[width=\linewidth]{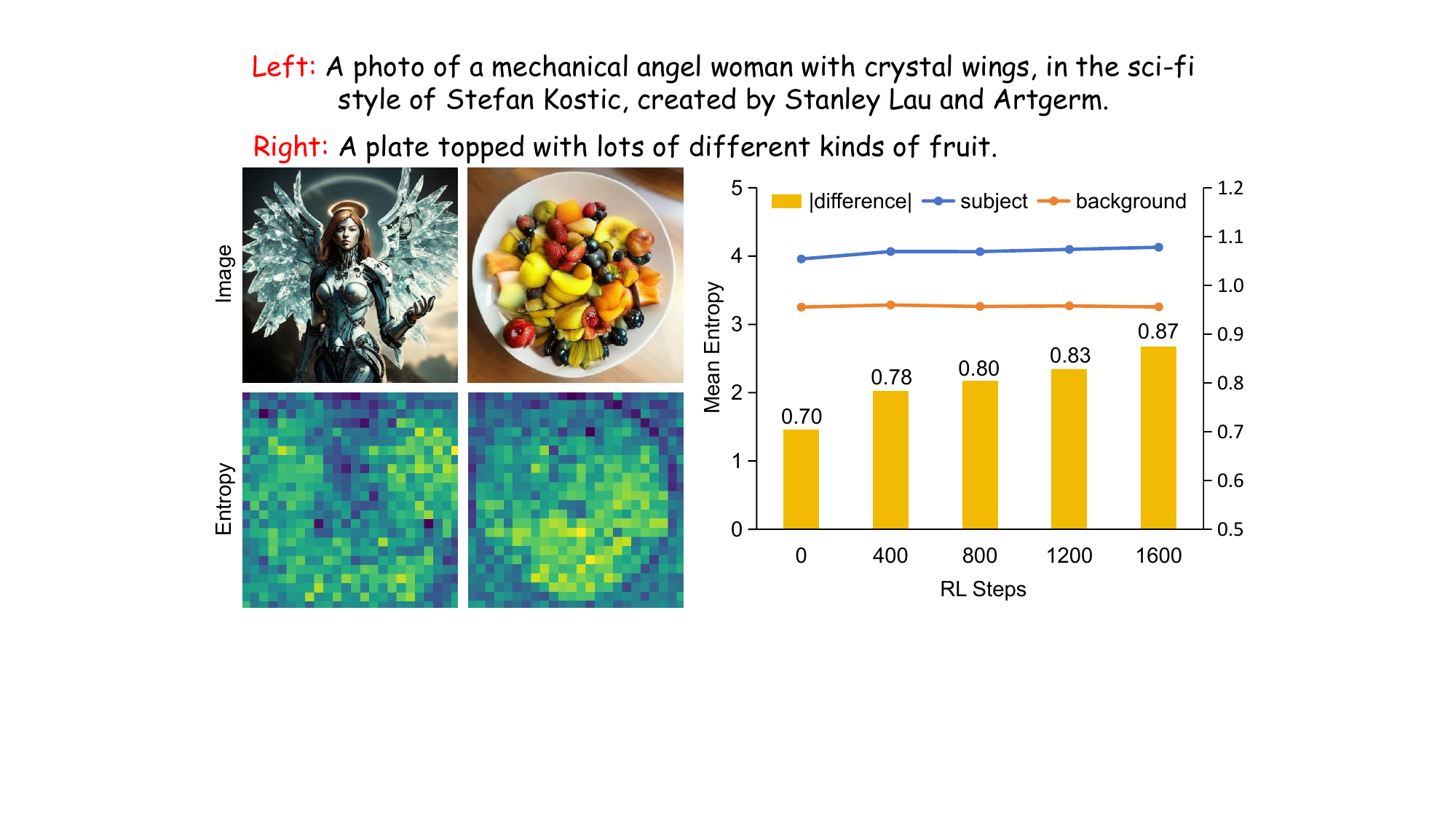}
    }
    \caption{In each figure, the blue line represents the average entropy of the subject regions, while the orange line represents the average entropy of the background regions. $|\text{difference}|$ represents the average entropy difference between subject and background regions.}
    \label{fig:entropy_low_high}
\end{figure}

In section~\ref{sec:Image Spatial Structure}, we discuss the entropy distribution of images in AR generation. We observe that the entropy distribution of image tokens exhibits a spatial pattern and fails to maintain consistency in different prompts. We further provide some examples and statistical results to illustrate this point, as shown in Fig.~\ref{fig:entropy_low_high}

We argue that this entropy distribution is directly related to the complexity of the prompts. Specifically, for composition prompts, such as Geneval-style prompts, the entropy of the subject region is lower than that of the background region (see Fig.~\ref{fig:entropy_low:a}). These prompts are relatively simple, including only the necessary keywords without excessive modifiers. As a result, the model has higher uncertainty when generating tokens in the background region due to the lack of sufficient prompt information. For complex prompts, such as the HPS testing data, the subject regions have a higher information density, resulting in higher entropy for tokens in these regions (see Fig.~\ref{fig:entropy_low:b}).

Based on this insight, we conduct corresponding experiments to verify this point. We analyze the images generated by Janus-Pro and the corresponding entropy distribution, where prompts come from the Geneval Benchmark and HPS testing data. We use RMBG v2.0~\citep{RMBG}, a state-of-the-art background removal model, to separate the foreground and background regions of the generated images. Then, we calculate the average token entropy within the foreground region, as shown in Fig.~\ref{fig:entropy_low_high}. The statistical results further verify our view.

In addition, we further investigate the evolution of entropy distribution during the RL training process. We find that RL training does not alter the original entropy pattern, but instead further reinforces it, which is similar to observations in LLMs~\citep{li2025llms, vassoyan2025ignore}. This phenomenon means that the entropy difference between the subject and background regions gradually increases (see the histogram in Fig.~\ref{fig:entropy_low_high}), resulting in more pronounced entropy (information) changes for high-gradient tokens in their neighborhood. Such information variation and complex structural regions are simultaneously coupled to these tokens, prompting us to include them in the scope of critical tokens for focused optimization.

\section{More Comparison Results with related work}
In the field of AR image generation based on RLVR, \citep{jiang2025t2i} and \citep{pan2025focusdiff} are representative works that are close to ours. The differences are that \citep{jiang2025t2i} introduces semantic-CoT before image generation and uses a combination of multiple reward models, including HPS~\citep{wu2023human}, GroundingDINO~\citep{liu2024grounding}, GIT~\citep{wang2022git}, and LLaVA-OneVision-7B~\citep{li2024llava}, and constructs training data that includes both Geneval-style and T2i-compbench-style prompts. We only utilize Geneval rewards and Geneval-style data. \citep{pan2025focusdiff} constructs fine-grained paired prompt-image training data and first trains the model on paired data with images, then operates GRPO training without images. In contrast, we only perform image-free RL training. Comparison results on Geneval and T2i-compbench are shown in the Table~\ref{tab:appendix_t2i_compbench}. Our method achieves a significant lead on Geneval and also obtains comparable results on T2i-compbench.

\label{sec:More Comparison Results with related work}
\begin{table}[ht]
\centering
\caption{ \label{tab:appendix_t2i_compbench} Comparison results with T2I-R1 and Focus-Diff on Geneval and T2I-CompBench.}
\resizebox{0.99\linewidth}{!}{
\begin{tabular}{l c ccccccc}
\toprule
\multirow{2}{*}{Method$\uparrow$} & \multirow{2}{*}{GenEval$\uparrow$} & \multicolumn{6}{c}{T2I-CompBench} \\
\cmidrule(lr){3-8}
 &  & Color$\uparrow$ & Shape$\uparrow$ & Texture$\uparrow$ & Spatial$\uparrow$ & Non-Spat.$\uparrow$ & Complex$\uparrow$ \\
\midrule
T2I-R1~\citep{jiang2025t2i}               &0.79& 0.8130 & 0.5852 & 0.7243 & 0.3378 & 0.3090 & 0.3993 \\
Focus-Diff~\citep{pan2025focusdiff}           &0.85& 0.7996 & 0.5748 & 0.7007 & 0.3789 & 0.3098 & 0.3912 \\
Janus-Pro-7B+GRPO    &0.87& 0.7478 & 0.3999 & 0.5849 & 0.2481 & 0.3090 & 0.3744 \\
Janus-Pro-7B+GCPO    &0.90& 0.7508 & 0.5173 & 0.7030 & 0.3824 & 0.3133 & 0.3888 \\
\bottomrule
\end{tabular}
}
\end{table}

\clearpage
\section{More Visual Comparison Results}
\label{sec:More Visual Comparison Result}
In this section, we further provide visual comparison results on the composition image task and image quality task to better illustrate the effectiveness and compatibility of our method. We will provide prompts for all images in Sec.~\ref{sec:Used prompts in this section}.

\subsection{Comparison on LlamaGen}
\begin{figure}[htbp]
    \centering
    \subcaptionbox{Visual comparison results on HPS benchmark.\label{fig:llamagen_hps:a}}[.48\linewidth]{
        \includegraphics[width=\linewidth]{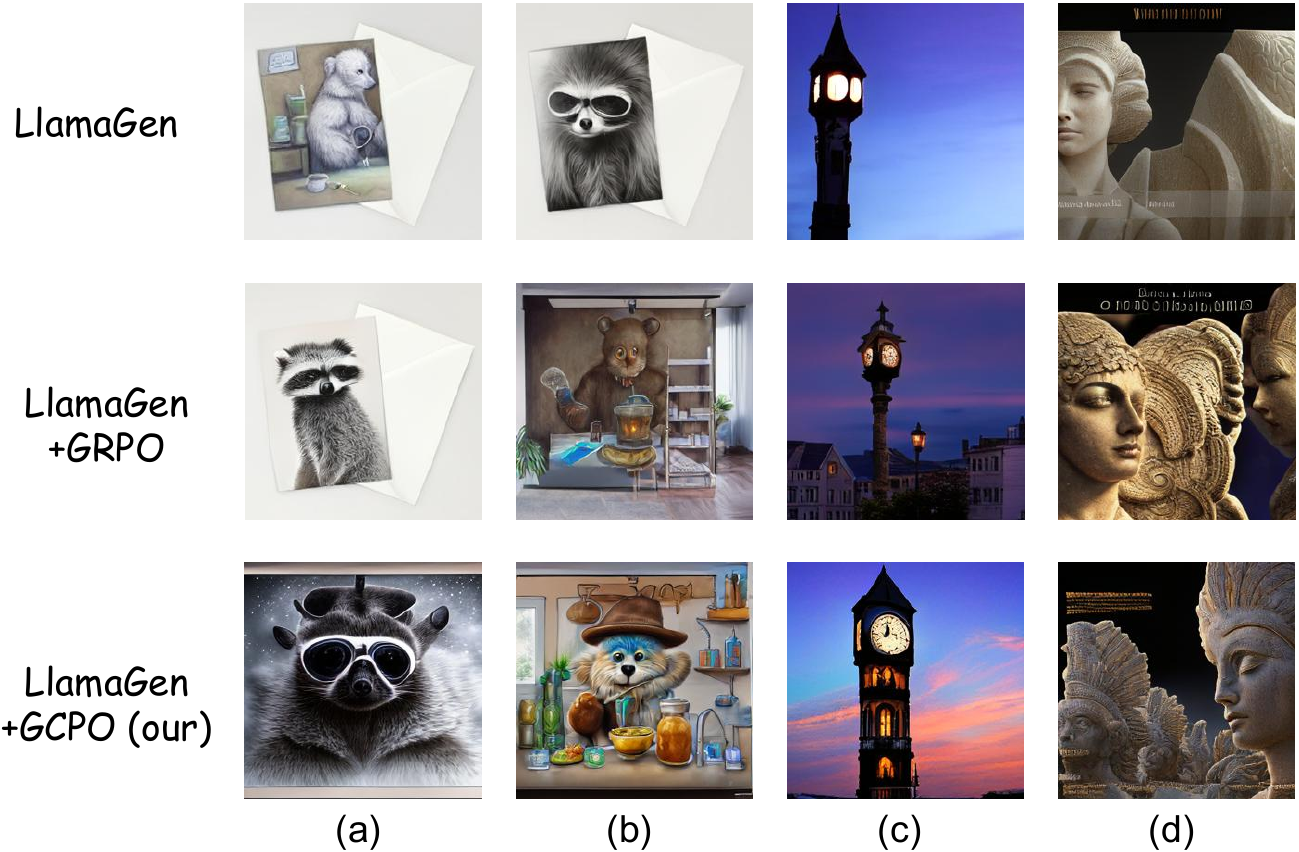}
    }
    \hfill
    \subcaptionbox{Visual comparison results on Geneval\label{fig:llamagen_geneval:b}}[.48\linewidth]{
        \includegraphics[width=\linewidth]{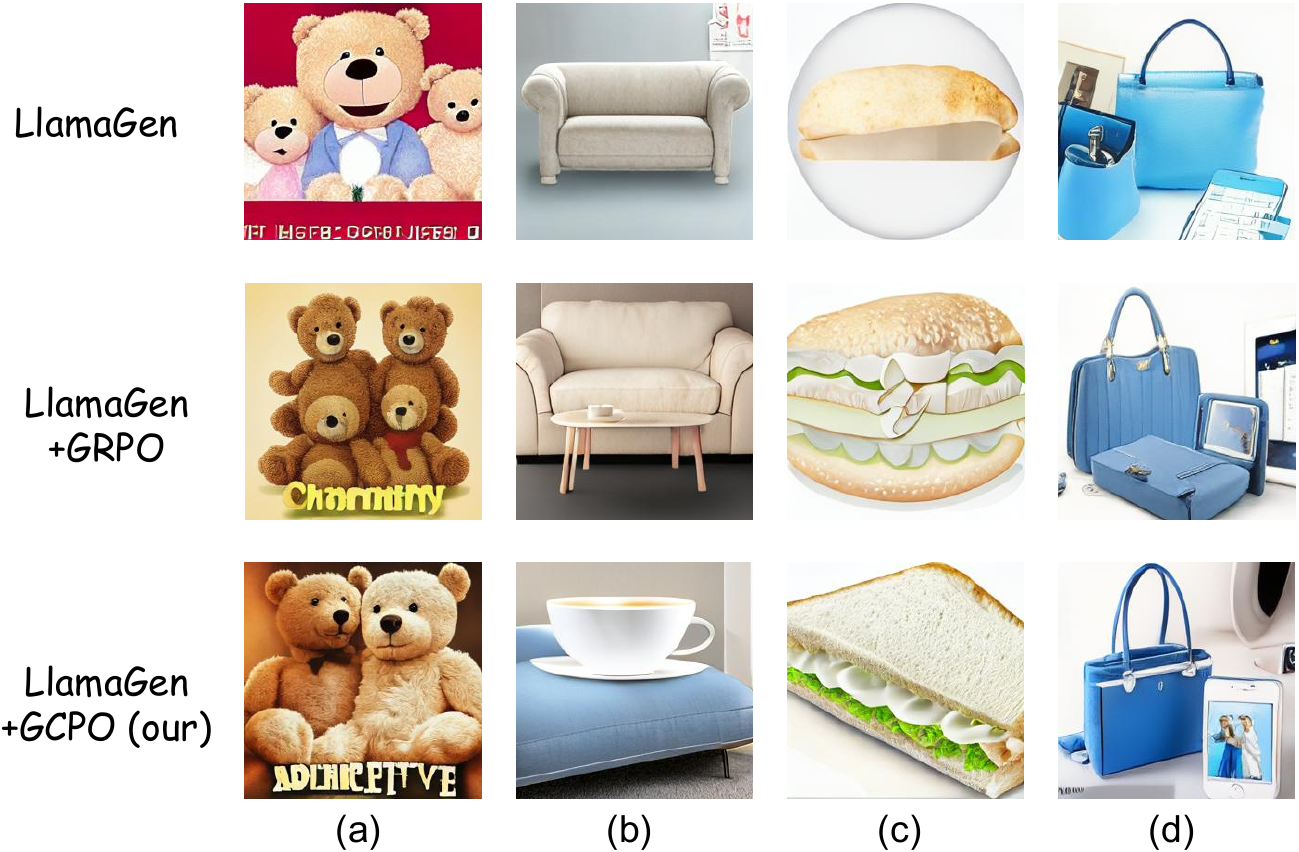}
    }
    \caption{Qualitative comparison between the LlamaGen + GRPO and LlamaGen + GCPO (our) trained with HPS reward. Prompts are in Sec.~\ref {sec:Used prompts in this section}
}
    \label{fig:llamagen_hps_geneval}
\end{figure}

We only use the HPS Reward on LlamaGen and abandon the Geneval reward. This is because we find that the Geneval score for each image generated by LlamaGen is close to 0 and does not provide effective training rewards. Nevertheless, the model trained with the HPS reward not only demonstrates better generation quality, as shown in~\ref{fig:llamagen_hps_geneval}, but also further improves performance on Geneval.

\subsection{Comparison on Janus-Pro 1B}
\begin{figure}[htbp]
    \centering
    \includegraphics[width=\linewidth]{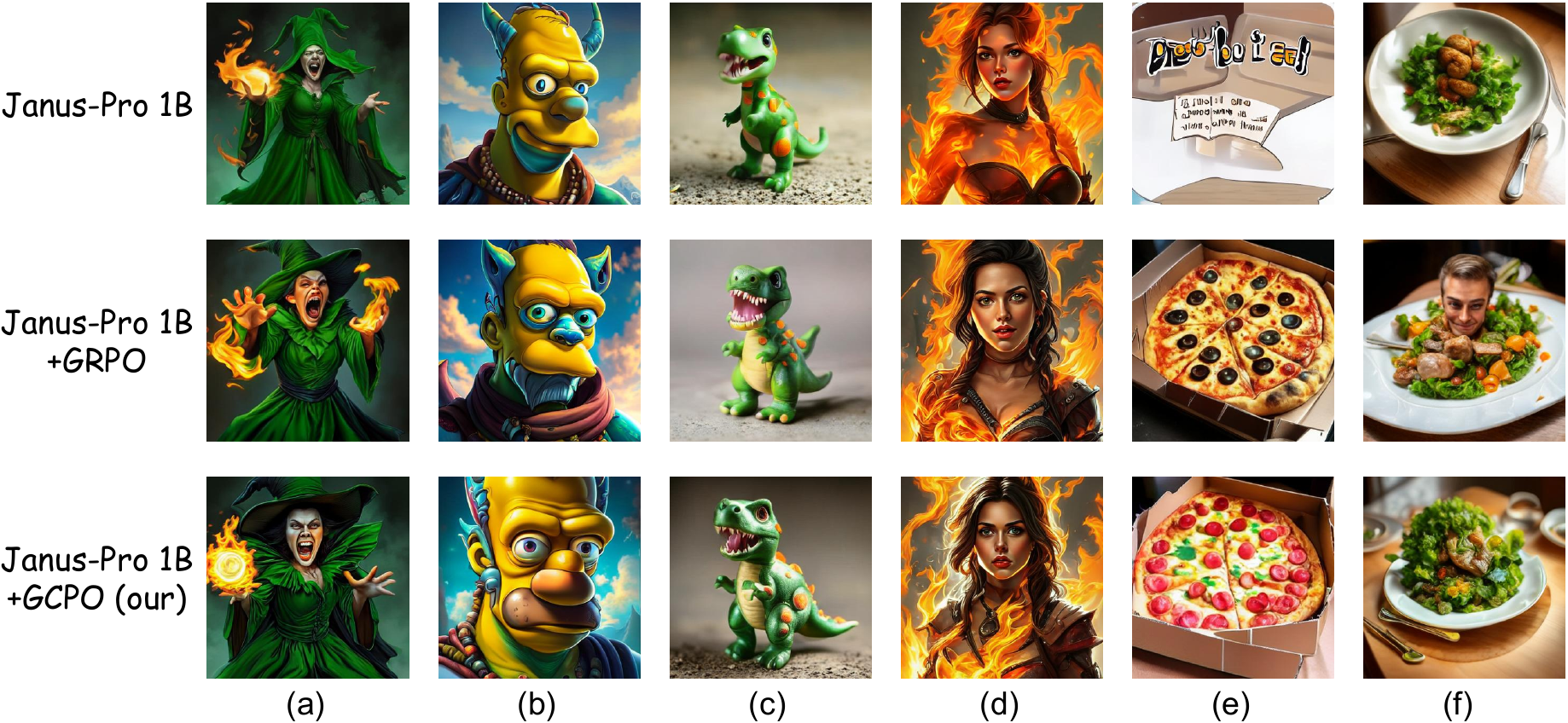}
    \caption{Qualitative comparison between the Janus-Pro 1B + GRPO and Janus-Pro 1B + GCPO (our) trained with HPS reward. Prompts are in Sec.~\ref {sec:Used prompts in this section}}
    \label{fig:1b_hps}
\end{figure}

\begin{figure}[htbp]
    \centering
    \includegraphics[width=\linewidth]{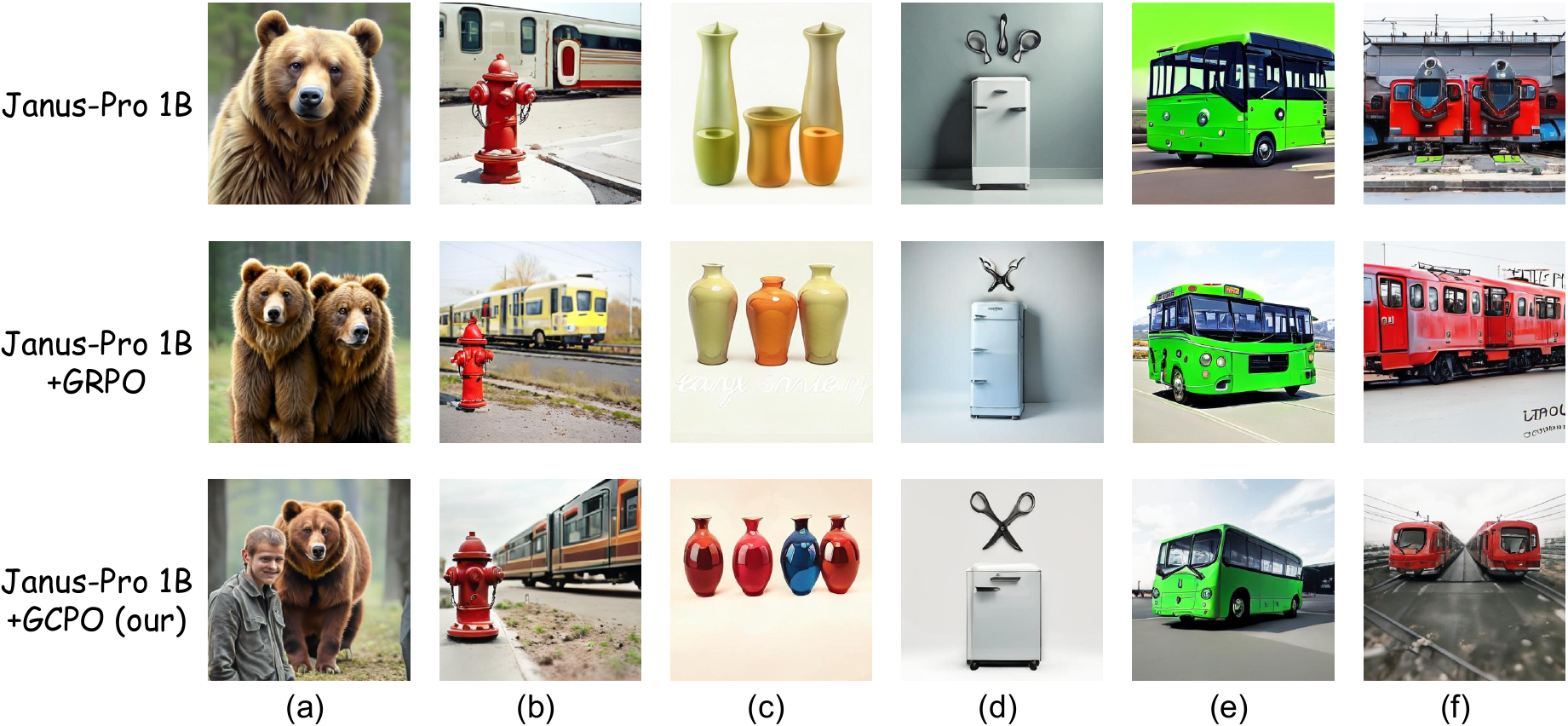}
    \caption{Qualitative comparison between the Janus-Pro 1B + GRPO and Janus-Pro 1B + GCPO (our) trained with Geneval reward. Prompts are in Sec.~\ref {sec:Used prompts in this section}}
    \label{fig:1b_geneval}
\end{figure}

We provide more qualitative comparison results on Janus-Pro 1B, as shown in~\ref{fig:1b_hps} (HPS reward) and~\ref{fig:1b_geneval} (Geneval reward). These visual results further demonstrate the effectiveness of our approach.

\subsection{More visual results during RLVR training}
\begin{figure}[htbp]
    \centering
    \includegraphics[width=0.9\linewidth]{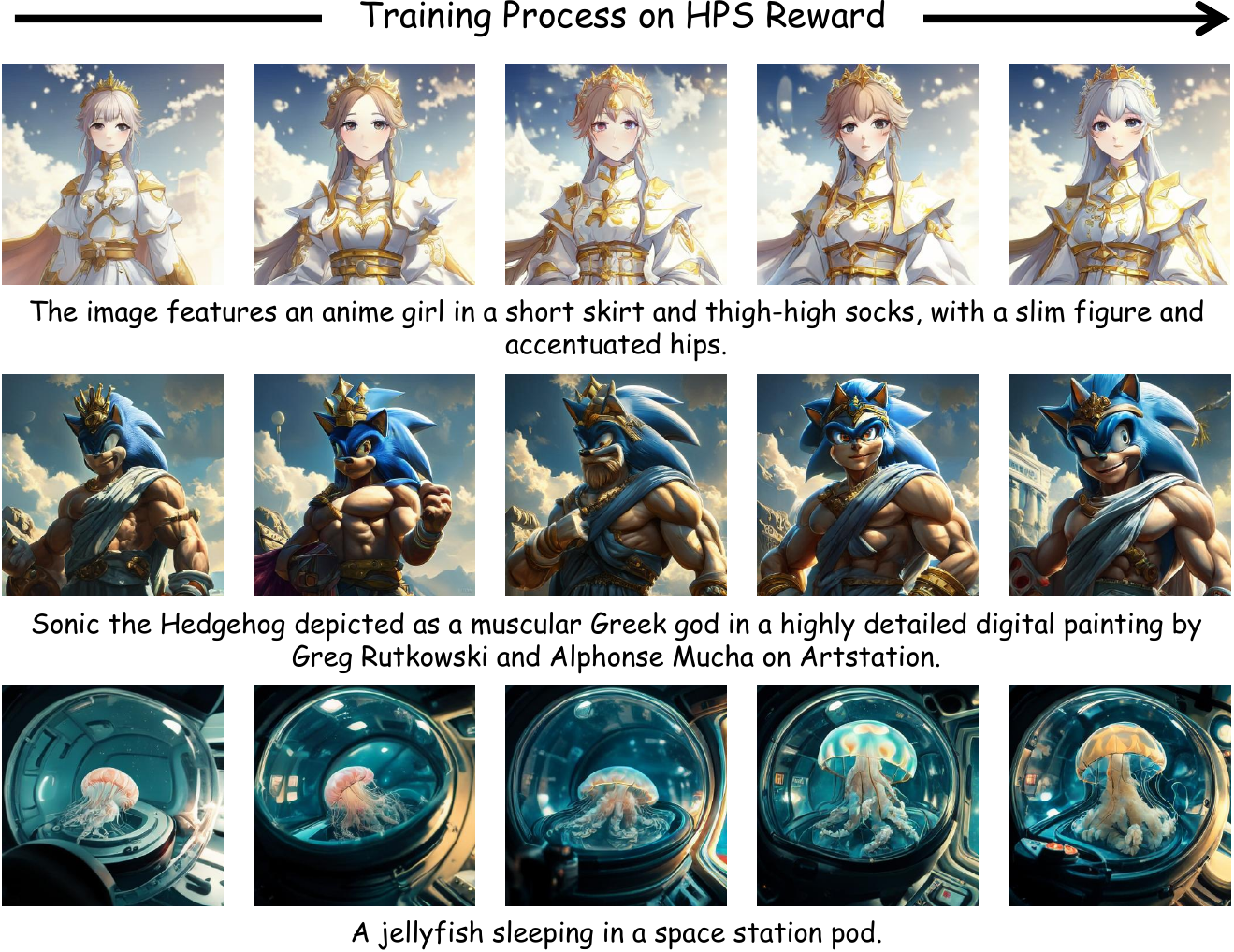}
    \caption{We visualize the generated samples during the optimization of Janus-Pro 7B trained with HPS reward. As training progressed, the model is steadily optimized towards the target of Human Preference Alignment.}
    \label{fig:7b_process_hps}
\end{figure}

To better understand the training dynamics of our method, we visualize the results of samples generated by the same prompts during training, as shown in Fig.~\ref{fig:7b_process_hps}. These qualitative results intuitively demonstrate how the model continuously optimizes towards the goal of improving image quality and Human Preference Alignment as training progresses.

\subsection{Used prompts in this section}
\label{sec:Used prompts in this section}
The prompts used in Fig.~\ref{fig:llamagen_hps:a} are:
\begin{enumerate}[label=(\alph*)]
    \item The image is of a raccoon wearing a Peaky Blinders hat, surrounded by swirling mist and rendered with fine detail.
    \item A teddy bear mad scientist mixing chemicals depicted in oil painting style as a fantasy concept art piece.
    \item A clock tower with lighted clock faces, against a twilight sky.
    \item The image features a closeup portrait of stone angel statues, created with the Unreal Engine and featuring intricate details by various artists.
\end{enumerate}

The prompts used in Fig.~\ref{fig:llamagen_geneval:b} are:
\begin{enumerate}[label=(\alph*)]
    \item a photo of two teddy bears
    \item a photo of a couch below a cup
    \item a photo of a white sandwich
    \item a photo of a blue handbag and a white cell phone
\end{enumerate}
The prompts used in Fig.~\ref{fig:1b_hps} are:
\begin{enumerate}[label=(\alph*)]
    \item Wicked witch casting fireball dressed in green with screaming expression.
    \item The image is a portrait of Homer Simpson as a Na'vi from Avatar, created with vibrant colors and highly detailed in a cinematic style reminiscent of romanticism by Eugene de Blaas and Ross Tran, available on Artstation with credits to Greg Rutkowski.
    \item A small green dinosaur toy with orange spots standing on its hind legs and roaring with its mouth open.
    \item Mila Kunis portrayed as a fire elemental in a highly detailed digital painting.
    \item A pizza is displayed inside a pizza box.
    \item A portrait of a dinner dish of a protein and greens.
\end{enumerate}

The prompts used in Fig.~\ref{fig:1b_geneval} are:
\begin{enumerate}[label=(\alph*)]
    \item a photo of a person and a bear
    \item a photo of a fire hydrant and a train
    \item a photo of four bowls
    \item a photo of a baseball glove right of a bear
    \item a photo of a green bus
    \item a photo of two trains
\end{enumerate}

\end{document}